\documentclass[final,3p,times,twocolumn]{elsarticle}

\usepackage{amsmath,amssymb}
\usepackage{pdflscape}
\usepackage{booktabs}
\usepackage{tabularx}
\usepackage{array}
\usepackage{graphicx}
\usepackage{microtype}
\usepackage{url}
\usepackage{float}
\usepackage[section]{placeins}
\begin{document}

\begin{frontmatter}

\title{A Compact Stance-Indexed Anterior--Posterior COP Representation for Parkinson's Disease Classification from Plantar VGRF}

\author[aff1]{Md. Sifat}
\ead{mdsifat371@gmail.com}

\author[aff1]{Sania Akter}
\ead{saniaaker@gmail.com}

\author[aff1]{Akif Islam}
\ead{iamakifislam@gmail.com}

\author[aff1]{Md. Ekramul Hamid}
\ead{ekram_hamid@ru.ac.bd}

\affiliation[aff1]{
    organization={Department of Computer Science and Engineering, University of Rajshahi},
    city={Rajshahi},
    country={Bangladesh}
}

\begin{abstract}
Parkinson's disease alters gait and bilateral coordination, but machine-learning performance also depends on how continuous gait signals are represented. This study investigates whether preserving anterior--posterior center-of-pressure (AP-COP) information at fixed locations across normalized stance provides a compact and informative representation of plantar-force gait signals. Bilateral vertical ground reaction force recordings from 165 participants in the Gait in Parkinson's Disease Database were evaluated using repeated fully nested participant-level cross-validation. We propose AP-COP10, comprising AP-COP position and bilateral asymmetry across five stance windows. AP-COP10 achieved an AUC of 0.894 and outperformed three harmonized literature-derived COP representations under the same evaluation pipeline. The complementary 25 non-AP-COP descriptors alone achieved an AUC of 0.856, while the complete 35-feature representation achieved 0.908. Removing AP-COP10 from the complete representation produced a statistically supported loss in discrimination, whereas adding the complementary descriptors to AP-COP10 yielded only a small, unsupported improvement. Feature competition indicated that the most informative stance-indexed descriptors were concentrated in early and early-mid stance, while source-study holdout and sensor-perturbation analyses supported the robustness of the representation. These findings indicate that stance-indexed AP-COP retains discriminative information that is not readily recovered by broader engineered gait descriptors, supporting compact and interpretable representations for machine-learning analysis of pathological gait.
\end{abstract}


\end{frontmatter}

\section{Introduction}
\label{sec:introduction}

In machine-learning analysis of gait, predictive performance depends not only on the choice of classifier but also on how continuous sensor signals are represented. This issue is particularly relevant in Parkinson's disease (PD), where gait impairment involves changes in pace, rhythmicity, postural stability, variability, and bilateral coordination \cite{mirelman2019,bouca2020,bloem2021}. Quantitative gait measures are increasingly being investigated as digital biomarkers for characterizing mobility impairment and monitoring disease-related changes \cite{russo2025,mancini2025}. Plantar-force sensing is attractive in this context because instrumented insoles provide direct measurements of foot--ground interaction during walking and support the extraction of temporal, kinetic, and spatial gait information \cite{zhang2023insole,wang2026survey}. The Gait in Parkinson's Disease Database (GaitPDB), which contains bilateral plantar vertical ground reaction force (VGRF) recordings from participants with PD and healthy controls across three source studies, has therefore become a widely used resource for computational gait analysis \cite{hausdorff2008}.

Among the quantities derived from plantar-force measurements, the center of pressure (COP) provides an interpretable description of how the resultant plantar load progresses beneath the foot during stance \cite{han1999,lugade2014}. Altered and asymmetric COP behavior has been reported in PD \cite{shin2020,zhang2024}, supporting its relevance for quantitative gait characterization. COP is fundamentally a continuous trajectory, however, and its information content depends on how that trajectory is represented. Global statistics, geometric measures, bilateral summaries, and dynamic descriptors compress different aspects of the underlying signal, and whole-trajectory summaries do not necessarily preserve where within stance a discriminative difference occurs. Continuous biomechanical analyses have similarly shown that localized effects may be obscured by scalar reduction \cite{pataky2014}.

This creates a representation-level problem. Existing COP-based studies differ in feature construction, participant selection, prediction task, model choice, and validation procedure, making their reported performance difficult to interpret as evidence that one representation is intrinsically more informative than another. More fundamentally, it remains unclear whether discriminative COP information is distributed broadly across the trajectory or concentrated within particular portions of stance that may be attenuated when the signal is reduced to whole-trajectory summaries. A controlled comparison under a common evaluation framework is therefore needed to isolate the effect of representation content.

Normalized stance provides a useful middle ground between aggressive whole-trajectory compression and high-dimensional waveform modeling. COP progression is known to vary systematically across different portions of stance and is influenced by walking speed, foot position, and foot structure \cite{chiu2013speed,lugade2014,buldt2018}. Retaining COP information at corresponding locations of normalized stance can therefore preserve the location of biomechanical information while remaining low dimensional and interpretable. This motivates preserving localized anterior--posterior COP information across normalized stance rather than summarizing the trajectory only at the whole-stance level. In the present work, five fixed windows are used to derive AP-COP position and corresponding bilateral asymmetry, yielding a compact 10-candidate representation termed AP-COP10.

We evaluate AP-COP10 under a common participant-level repeated fully nested framework against harmonized literature-derived COP representations and a broader 35-candidate engineered feature space spanning temporal-rhythm, dynamic-support, bilateral-asymmetry, COP-rollover, and vertical-loading descriptors. To examine how the stance-indexed descriptors contribute within this broader representation, the remaining 25 non-AP-COP descriptors are also evaluated independently. Together, these comparisons assess the discriminative information retained by the compact stance-indexed representation, by the complementary engineered descriptors, and by their combined feature space.

The main contributions are:

\begin{enumerate}
    \item A compact stance-indexed representation of anterior--posterior COP position and bilateral asymmetry across normalized stance.

    \item A controlled representation-level evaluation framework that isolates the effect of COP feature design from differences in experimental pipelines.

    \item An analysis of the discriminative value, compactness, and robustness of localized COP information for participant-level PD classification.
\end{enumerate}
\section{Related Work}
\label{sec:related}

\subsection{Plantar-force and COP analysis in Parkinsonian gait}

Plantar force measurements have long been used to characterize gait abnormalities in Parkinson's disease (PD), including altered force distribution and gait variability \cite{nieuwboer1999,hausdorff1998}. Subsequent reviews have documented broader spatiotemporal and biomechanical abnormalities in PD gait \cite{bouca2020,russo2025}. The center of pressure (COP) provides an interpretable spatial description of how the resultant load progresses beneath the foot during stance. COP trajectories have been characterized using in-shoe force measurements and normalized stance coordinates \cite{han1999,lugade2014}, with progression influenced by factors including walking speed, age, sex, and foot structure \cite{chiu2013speed,chiu2013elderly,buldt2018}. Altered and asymmetric COP behavior has also been reported in PD \cite{shin2020,zhang2024}.

More recent COP, plantar-pressure, and wearable-insole studies have expanded analysis to regional pressure dynamics, postural-sway measures, and larger multidomain feature spaces \cite{fadil2021,sun2023,nanayakkara2025}. Together, these studies establish plantar force and COP as useful sources for Parkinsonian gait analysis while showing that the same underlying trajectory can be represented in substantially different ways.

\subsection{COP representations for machine-learning gait analysis}

Machine-learning studies have used several forms of COP representation. Global statistics such as mean and standard deviation provide compact summaries of overall COP position and dispersion \cite{alam2017}. Bilateral and geometric descriptors, including AP-COP dispersion and COP-path symmetry, retain information about inter-limb relationships and overall COP behavior \cite{jin2026}. Other approaches use richer dynamic or nonlinear descriptors, including COP velocity, acceleration, jerk, path efficiency, and entropy-related measures \cite{tong2021}.

Trajectory- and phase-resolved analyses provide a complementary approach by retaining information across the progression of stance \cite{pataky2014}. Biomechanical studies have shown that COP characteristics vary across normalized stance \cite{chiu2013speed,lugade2014,buldt2018}, suggesting that both COP position and its location within the stance cycle may be informative. Bilateral asymmetry provides another complementary property, as inter-limb differences in COP behavior have been reported in PD \cite{shin2020,jin2026}. Together, these studies indicate that both AP-COP behavior and its localization within stance may be informative, while existing representations differ substantially in the trajectory information they preserve.

Existing studies, however, use different feature definitions and experimental pipelines, making their reported performance difficult to compare directly. A controlled representation-level comparison can therefore provide a clearer assessment of the information retained by different COP encodings.

\subsection{Machine learning on the Gait in Parkinson's Disease Database}

GaitPDB has also been studied using a broad range of machine-learning approaches beyond COP-specific features. Earlier work used engineered VGRF, temporal, and frequency-domain descriptors with classical classifiers \cite{zeng2016,khoury2019,farashi2020}, while other studies focused on stance-related force characteristics or disease-stage classification \cite{balaji2020,farashi2021}. More recent work has increasingly used deep learning, including convolutional models, explainable architectures, and graph-based approaches for multichannel plantar-force data \cite{elmaachi2020,choi2026,wang2026graph}.

These studies demonstrate the predictive richness of GaitPDB, but their different representations, model designs, and validation strategies make published performance difficult to interpret as evidence for the superiority of one representation. The present study therefore focuses on the controlled comparison of interpretable COP representations under a common participant-level framework, with particular attention to the information preserved by stance-indexed AP-COP.
\section{Materials and Methods}
\label{sec:methods}

\subsection{Study Design and Dataset}
\label{sec:study_design}

\begin{figure*}[t]
\centering
\includegraphics[width=0.98\textwidth]{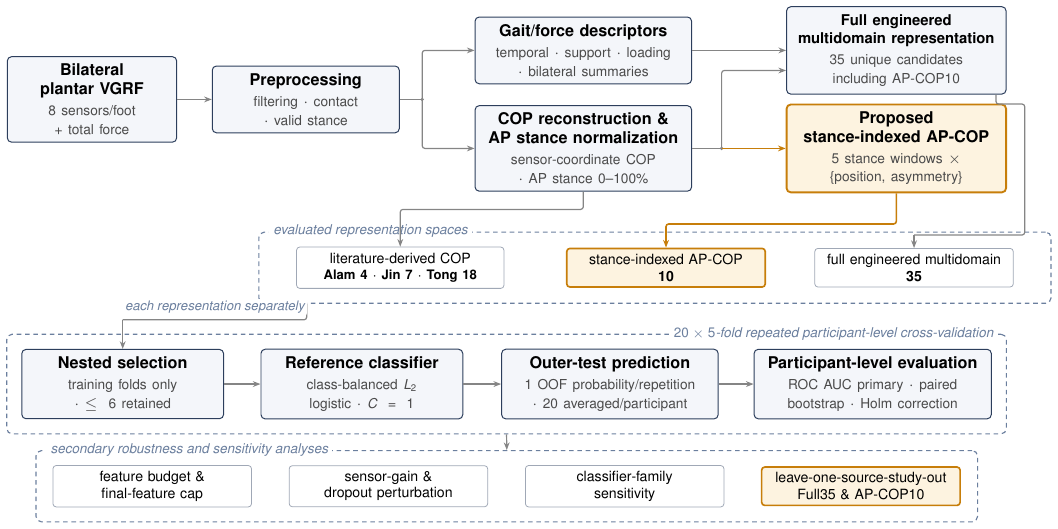}
\caption{Study workflow from bilateral plantar VGRF preprocessing and feature construction through fully nested participant-level evaluation.}
\label{fig:workflow}
\end{figure*}

The study was designed to isolate the effect of representation content on
participant-level Parkinson's disease classification. Three representation
groups were evaluated under a common machine-learning pipeline: the proposed
10-candidate stance-indexed AP-COP representation (AP-COP10), three
literature-derived COP representations, and a complete 35-candidate engineered
multidomain representation (Full35). The 25 non-AP-COP descriptors contained
within Full35 were also evaluated independently. All representations used the
same participants, signal preprocessing, outer data partitions, training-only
feature selection, and final classifier. This design allows differences in
predictive performance to be interpreted primarily in terms of representation
content rather than differences in the surrounding experimental pipeline.
Figure~\ref{fig:workflow} summarizes the study workflow.

The proposed representation was evaluated in two complementary comparisons.
First, it was compared with COP descriptors derived from Alam et al.
\cite{alam2017}, Jin \cite{jin2026}, and Tong et al. \cite{tong2021}, each
harmonized to the common participant-level pipeline. Second, AP-COP10 was
evaluated as a strict subset of Full35, allowing the contribution of additional
temporal, support, bilateral, COP-rollover, and vertical-loading descriptors to
be examined without changing the evaluation framework. The remaining 25
descriptors were additionally evaluated on their own to assess their predictive
information in the absence of stance-indexed AP-COP.

\subsubsection{Dataset and participant-level prediction task}

Experiments used GaitPDB from PhysioNet \cite{hausdorff2008}. The database
provides bilateral plantar VGRF signals from eight individual force sensors
beneath each foot plus left and right total-force channels, sampled at
100~Hz. The database documentation lists 93 participants with idiopathic PD
and 73 healthy controls collected across three studies. In the present
analysis, one \texttt{\_01} usual-walking recording was retained per
participant from the Ga, Ju, and Si source studies. One control listed in
the demographic table lacked a matching \texttt{\_01} signal file, leaving
an analytic cohort of 165 participants: 93 PD and 72 controls.

Source-study counts were 47 for Ga, 54 for Ju, and 64 for Si. Participant
characteristics are shown in Table~\ref{tab:participants}. The participant
was the prediction and analysis unit. Individual samples, gait cycles,
stances, windows, and repeated steps were used only to construct
participant-level descriptors and were never treated as independent
machine-learning observations. This participant-level design prevents
repeated measurements from the same individual from contributing
independent observations to model evaluation.

\begin{table}[H]
\centering
\caption{Participant characteristics by source study and diagnosis. Continuous
variables are median [IQR].}
\label{tab:participants}
\scriptsize
\setlength{\tabcolsep}{3.1pt}
\renewcommand{\arraystretch}{1.04}
\resizebox{0.98\columnwidth}{!}{%
\begin{tabular}{llrrrrr}
\toprule
\textbf{Study} & \textbf{Group} & \textbf{$n$} & \textbf{Age (y)} &
\textbf{Female} & \textbf{Weight (kg)} & \textbf{Speed (m/s)} \\
\midrule
Ga & Control & 18 & 69 [67--77] & 44\% & 70 [64--82] & 1.2 [1.1--1.3] \\
Ga & PD      & 29 & 71 [68--78] & 31\% & 74 [62--84] & 1.0 [0.8--1.1] \\
Ju & Control & 25 & 65 [60--68] & 52\% & 72 [60--80] & 1.2 [1.1--1.3] \\
Ju & PD      & 29 & 68 [64--74] & 45\% & 68 [65--80] & 1.1 [0.9--1.2] \\
Si & Control & 29 & 57 [53--62] & 38\% & 74 [67--83] & 1.3 [1.1--1.4] \\
Si & PD      & 35 & 63 [56--68] & 37\% & 74 [66--80] & 1.1 [1.0--1.2] \\
\bottomrule
\end{tabular}%
}
\end{table}

\subsection{Signal Processing and Feature Construction}

\subsubsection{Signal preprocessing and stance extraction}

The first and final 10~s of each recording were removed. Left and right
total VGRF were clipped at zero and median filtered using a five-sample
window. Foot contact was detected independently with a 20-N threshold.
Contact interruptions lasting at most five samples were closed, runs
shorter than 20 samples were removed, and candidate stances were restricted
to 0.30--2.00~s. Participants were required to have at least eight valid
stances for each foot. The retained left- and right-foot stance runs were
ordered chronologically, and one boundary stance was removed from each end
of the active walking interval.

For timing and multidomain descriptors, contact onset defined the step
event. Consecutive events separated by less than 0.20~s were excluded.
Alternating-foot step intervals were restricted to 0.25--1.50~s; same-foot
stride, stance, and swing durations were restricted to 0.50--2.50,
0.30--2.00, and 0.05--1.50~s, respectively. Body force for normalization of
vertical-loading descriptors was estimated as the median bilateral total
force among samples where combined force exceeded 100~N.

\subsubsection{Sensor-coordinate COP reconstruction and AP stance normalization}
\label{subsec:cop}

Negative individual-sensor force values were clipped to zero. AP-COP was
derived independently for each foot using fixed anterior--posterior sensor
coordinates
\[
s_j=[-800,-400,-400,0,0,400,400,800].
\]
For sensor force $f_j(t)$, the raw force-weighted AP load-center coordinate
was
\begin{equation}
q_{\mathrm{raw}}(t)=
\frac{\sum_{j=1}^{8}f_j(t)s_j}{\sum_{j=1}^{8}f_j(t)},
\label{eq:qraw}
\end{equation}
when the summed individual-sensor force exceeded $10^{-8}$. The resulting
coordinate is scaled to $[0,1]$,
\begin{equation}
q(t)=\frac{q_{\mathrm{raw}}(t)+800}{1600},
\label{eq:qnorm}
\end{equation}
and is interpreted as a sensor-coordinate AP load-center proxy rather than
an anatomical COP distance, because the underlying coordinates represent
discrete sensor locations rather than measured anatomical distances.

Within each detected stance, non-finite AP-COP values were removed. A
stance was discarded from AP-COP analysis when fewer than 30 finite
samples remained. Accepted AP-COP trajectories were linearly resampled to
101 points corresponding to 0--100\% normalized stance and smoothed using
an 11-point, second-order Savitzky--Golay filter with interpolation-based
edge handling \cite{savitzky1964}.

For literature-derived descriptors requiring two-dimensional COP, the same
force-weighted reconstruction was applied using the sensor-plane
coordinates adopted for the corresponding reference representations, and
the resulting ML and AP trajectories were summarized under the harmonized
stance extraction described above. This common preprocessing was
intentional: the goal was to compare representation content rather than
reproduce the original studies' entire signal-processing pipelines.

\subsubsection{Proposed stance-indexed AP-COP representation}
\label{subsec:proposed}

The proposed representation preserves AP-COP position and bilateral
asymmetry at five fixed windows distributed across normalized stance:
\[
W=\{5\text{--}15,\;25\text{--}35,\;45\text{--}55,\;65\text{--}75,\;85
\text{--}95\}\%.
\]
The windows specify \emph{when} within normalized stance AP-COP is
summarized rather than fixed physical locations on the foot. They were
fixed before the reported classification runs and were not selected by
searching PD--control results. The windows provide approximately uniform
coverage from early to late stance without concentrating all descriptors
in a single phase.

For participant $i$, foot $F\in\{L,R\}$, stance $r$, and window $w=[a,b]$,
the stance-level AP-COP position was
\begin{equation}
m_{i,F,r}^{(w)}=
\frac{1}{b-a+1}\sum_{k=a}^{b}q_{i,F,r}(k),
\label{eq:windowmean}
\end{equation}
and repeated stances were summarized separately for the two feet:
\begin{equation}
Q_{i,F}^{(w)}=\operatorname{median}_{r}m_{i,F,r}^{(w)}.
\label{eq:footmedian}
\end{equation}
The participant-level bilateral position descriptor was
\begin{equation}
P_i^{(w)}=\frac{Q_{i,L}^{(w)}+Q_{i,R}^{(w)}}{2}.
\label{eq:bilatpos}
\end{equation}
To quantify left--right differences on a relative scale, we used a
Robinson-type symmetry index based on the absolute bilateral difference
normalized by the mean magnitude of the two side-specific coordinates, a
widely used class of discrete gait-symmetry measures
\cite{robinson1987,viteckova2018}. The corresponding bilateral asymmetry
descriptor was
\begin{equation}
A_i^{(w)}=
\frac{\left|Q_{i,L}^{(w)}-Q_{i,R}^{(w)}\right|}
{0.5\left(\left|Q_{i,L}^{(w)}\right|+\left|Q_{i,R}^{(w)}\right|\right)}.
\label{eq:asym}
\end{equation}
The primary asymmetry formulation expresses bilateral AP-COP separation
relative to the mean magnitude of the two side-specific coordinates,
providing a dimensionless measure that can be compared across stance
windows. Because $q(t)$ is expressed on a heel-anchored $[0,1]$ scale, this
normalization depends on the chosen coordinate origin. Two origin-invariant
alternatives were therefore evaluated as sensitivity variants: the absolute
bilateral difference $\left|Q_{i,L}^{(w)}-Q_{i,R}^{(w)}\right|$, already
expressed in normalized AP-coordinate units, and the same difference
divided by the participant-level AP range.

Both foot summaries were finite for all participants and windows in the
reported analysis. Repeating the procedure over five windows yielded five
position and five asymmetry descriptors, for 10 participant-level
candidates. Figure~\ref{fig:aggregation} illustrates the construction. The
exact feature identifiers are provided in Supplementary Table~S1.

\begin{figure*}[t]
\centering
\includegraphics[width=0.94\textwidth]
{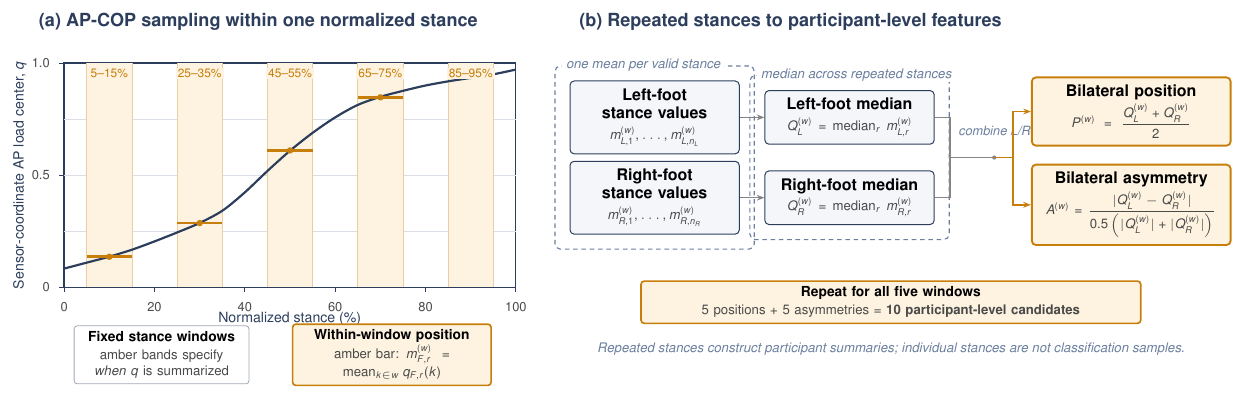}
\caption{Construction of the stance-indexed AP-COP representation. (a) Each
accepted AP-COP stance trajectory is normalized to 0--100\% stance and
sampled within five fixed windows. (b) Window values are aggregated across
repeated stances separately for the two feet, then combined into one
bilateral position and one bilateral asymmetry descriptor per window. The
five windows therefore yield 10 participant-level candidate features.}
\label{fig:aggregation}
\end{figure*}

\subsubsection{Complete engineered multidomain representation}
\label{subsec:full35}

The complete engineered representation (Full35) contained 35 unique
participant-level candidates spanning six descriptor families: six
temporal-rhythm descriptors, four dynamic-support descriptors, four
bilateral asymmetry descriptors outside the AP-COP family, six broader
COP-rollover descriptors (AP range, path length, median COP speed,
COP-speed coefficient of variation, monotonicity, and early slope), five
vertical-loading descriptors, and the 10 stance-indexed AP-COP descriptors
defined in Section~\ref{subsec:proposed}. Candidate definitions are
summarized in Table~\ref{tab:full35_families}; the complete feature
dictionary is provided in Supplementary Table~S2.

Full35 denotes a candidate feature space rather than a 35-variable fitted
model and contains AP-COP10 as a strict subset. The complementary
representation, termed \emph{Non-AP-COP25}, was obtained by removing the
10 stance-indexed AP-COP candidates from Full35. It therefore retained six
temporal-rhythm, four dynamic-support, four non-AP-COP bilateral-asymmetry,
six broader COP-rollover, and five vertical-loading descriptors. The same
training-only feature-selection procedure and six-feature maximum were
applied to Non-AP-COP25 as to the other representation spaces. Full35 and
Non-AP-COP25 therefore denote candidate pools rather than fixed
high-dimensional fitted models.

\begin{table}[t]
\centering
\caption{Descriptor families in the complete 35-candidate engineered
multidomain representation.}
\label{tab:full35_families}
\scriptsize
\setlength{\tabcolsep}{2.2pt}
\renewcommand{\arraystretch}{1.02}
\begin{tabularx}{\columnwidth}{p{1.75cm}p{0.5cm}X}
\toprule
\textbf{Family} & \textbf{$n$} & \textbf{Information retained} \\
\midrule
Temporal rhythm & 6 & Step/stride timing, stance and swing variability,
step-to-step irregularity. \\
Dynamic support & 4 & Double-support organization and stance-related
support structure. \\
Bilateral asymmetry & 4 & Normalized left--right differences in temporal
and loading summaries. \\
Broader COP rollover & 6 & AP range, path length, median speed, speed
variability, monotonicity, early slope. \\
Vertical loading & 5 & Normalized peak force, impulse, loading and
unloading behavior, waveform variability. \\
Stance-indexed AP-COP & 10 & Bilateral AP-COP position and asymmetry at
five fixed normalized-stance windows. \\
\midrule
\textbf{Total} & \textbf{35} & Complete engineered candidate pool. \\
\bottomrule
\end{tabularx}
\end{table}

\subsubsection{Literature-derived COP reference representations}
\label{subsec:litrefs}

All literature-derived sets were defined before their controlled runs and
were evaluated with the same downstream nested procedure as the proposed
representation. They should be interpreted as \emph{harmonized
reimplementations of published COP descriptor definitions}, not as
reproductions of the original papers' complete classifiers.

\noindent\textbf{Alam-derived COP4.}\enspace
Alam et al. used mean and standard deviation of COP coordinates as
machine-learning features \cite{alam2017}. The direct reference therefore
contained four left-foot descriptors: mean ML-COP, SD ML-COP, mean AP-COP,
and SD AP-COP. Stance-level quantities were summarized under the common
participant aggregation. Because the proposed representation is bilateral,
an eight-candidate bilateral extension of the same descriptor types was
also evaluated as a sensitivity analysis to confirm that the primary
result was not an artifact of the single-foot formulation.

\noindent\textbf{Jin-derived COP7.}\enspace
The Jin-derived COP representation contained seven COP-specific
descriptors \cite{jin2026}: AP-COP SD for the left and right feet, ML-COP
SD for the left and right feet, two-dimensional COP path length for the
left and right feet, and bilateral COP-path symmetry. Forefoot-to-rearfoot
loading ratio was deliberately excluded because it is a regional
plantar-loading descriptor rather than a COP descriptor. This distinction
is important because Jin reported moderate-to-large group effects for
forefoot-to-rearfoot loading and identified it as an influential
gait-derived quantity in the broader analysis. The present Jin7 comparator
therefore isolates the COP dispersion/path/symmetry family by design and
should not be interpreted as Jin's complete gait feature set.

\noindent\textbf{Tong-derived COP18.}\enspace
Tong et al. described linear and nonlinear characteristics extracted from
COP trajectories \cite{tong2021}. The primary controlled Tong-derived set
contained nine reproducible descriptor types retained separately for the
left and right feet: RMS ML-COP, RMS AP-COP, RMS total COP coordinate
magnitude, RMS two-dimensional COP velocity, RMS acceleration, RMS jerk,
COP path efficiency, ML-COP sample entropy, and AP-COP sample entropy,
yielding 18 candidates. COP path efficiency was defined as direct
start-to-end COP displacement divided by actual COP path length during
stance.

Descriptors that could not be deterministically reconstructed from the
source description---COP-track intersection measures (CSIP/CISP), whose
intersection algorithm was not specified sufficiently for reliable
reimplementation---were excluded rather than approximated, and
source-specific downstream procedures such as persistent-homology
processing, oversampling, and SVM classification were omitted, as the
present comparison isolates COP descriptor representation under a common
downstream classifier. Sample entropy used embedding dimension $m=2$ and
tolerance $r=0.20\times\mathrm{SD}$, disclosed because the source article
does not specify this parameterization. Secondary sensitivities (a
nine-candidate bilateral summary, an AP-focused eight-candidate set, and a
generous axis-expanded 30-candidate set) are reported in Results. The
complete feature-level reproduction and harmonization record is provided
in Supplementary Table~S3.

\noindent\textbf{Literature-derived COP union.}\enspace
To evaluate whether combining descriptor families from more than one
source study captures additional information relative to any single
comparator, an 11-candidate union of non-duplicated Alam- and Jin-derived
COP descriptors was constructed. This set was assembled for the present
comparison and was not used verbatim in any single prior paper; it is
reported alongside, but separately from, the three primary literature
contrasts.

\subsection{Model Development and Evaluation}

\subsubsection{Fully nested feature selection and validation}
\label{subsec:nested}

The entire feature-selection and model-development process was nested
within the outer training data so that no held-out participant influenced
feature eligibility, feature ranking, hyperparameter selection, or model
fitting, avoiding the optimistic bias that arises when feature or model
selection is performed outside the validation loop
\cite{varma2006,cawley2010}. Inner folds were used exclusively for model
development, whereas outer folds were used only for performance
estimation. Identical outer partitions were applied to every
representation, enabling paired participant-level comparison.

All model development occurred within each outer-training partition.
Missing values were median-imputed and candidates standardized using
training-only statistics. Elastic-net logistic hyperparameters
\cite{zou2005} were selected by inner five-fold cross-validation
stratified jointly by diagnosis and source study. Nine $C$ values
logarithmically spaced from $10^{-2}$ to $10^{1}$ and
$\mathrm{l1\_ratio}\in\{0.25,0.50,0.75,1.00\}$ were evaluated using mean
inner ROC AUC, with effective ties favoring stronger regularization.

Using the selected hyperparameters, feature stability was estimated from
100 stratified subsamples containing 75\% of the outer-training
participants, following the general principle that subsampling-based
stability can make variable selection less dependent on a single fitted
sample \cite{meinshausen2010}. A candidate was retained when its
elastic-net coefficient was nonzero in at least 60\% of subsamples. If no
feature met that threshold, the most stable candidate was retained.
Redundancy was then pruned using training-only absolute Spearman
correlation: when two retained descriptors had $|\rho|\geq0.85$, the less
stable member was removed. Remaining candidates were ordered by stability
and at most six entered the final class-balanced L2-regularized logistic
regression classifier ($C=1$, \texttt{liblinear}). No outer-test
participant contributed to imputation, scaling, hyperparameter selection,
stability estimation, redundancy pruning, feature ordering, or fitting.

Performance was estimated using 20 repetitions of five-fold
participant-level cross-validation (100 outer folds). Outer partitions
were stratified jointly by diagnosis and source study, and the exact same
outer splits were used for every representation. Each participant
therefore received one out-of-fold probability per repetition; the 20
probabilities were averaged to obtain a final participant-level
prediction. ROC AUC was the primary metric. Balanced accuracy,
sensitivity, specificity, and Brier score were secondary metrics, with a
fixed probability threshold of 0.5 for threshold-based measures.

\subsection{Statistical Analysis and Robustness}

\subsubsection{Statistical analysis}

AUC confidence intervals were obtained from 5,000 participant-level
bootstrap resamples stratified by diagnosis. The three direct literature
comparisons---proposed vs Alam-derived COP4, Jin-derived COP7, and
Tong-derived COP18---used matched participant-level bootstrap resampling
within source-study-by-diagnosis strata. Two-sided $p$ values used a
plus-one correction. Family-wise error across this three-comparison
primary literature family was controlled using Holm's sequentially
rejective procedure \cite{holm1979}.

The nested contrast between the complete and compact representations,
comparisons involving Non-AP-COP25, the literature-derived COP union, and
all remaining comparisons were treated as secondary analyses and were not included in the primary multiplicity family. Comparisons
involving Non-AP-COP25 used the same matched participant-level bootstrap
procedure within source-study-by-diagnosis strata.

For stance-window characterization, Hedges' $g$, bootstrap confidence
intervals, direction-agnostic univariable AUC, and outer-fold retention
frequencies were examined. Additional sensitivity analyses assessed the
coordinate-origin dependence of the relative asymmetry metric, predictive
information beyond walking speed, feature-budget and final-cap compactness,
simulated sensor gain variation and dropout, classifier family, and
leave-one-source-study-out performance. These analyses were used to probe
robustness and interpretation rather than to redefine the proposed
10-candidate representation.

\subsubsection{Secondary robustness analyses}

Feature-budget analysis truncated the training-derived stability ordering
of the complete representation at maximum budgets $k\in\{1,\dots,6\}$ and
refitted the final model, so the analysis evaluates at most $k$
descriptors rather than independent exact-$k$ selection; $k=6$ reproduces
the primary complete-representation result by construction. Final-cap
sensitivity separately reran fold-specific stability selection with
maximum final-feature caps of 4, 6, 8, 10, and 12. Classifier-family
sensitivity evaluated the fold-specific selected descriptors with
radial-basis-function support vector machine, random forest, and histogram
gradient boosting alongside logistic regression. Sensor perturbation
applied independent multiplicative gains with mean 1 and standard
deviation 0.05, 0.10, or 0.20 to the 16 individual plantar sensor channels
before COP extraction, together with random single- and two-sensor
dropout, using five random realizations per condition; models were trained
only on clean outer-training data and evaluated on perturbed outer-test
features. Source-study robustness held out Ga, Ju, and Si in turn, with
all feature development restricted to the remaining cohorts, and was
applied to both the complete and the compact representation.

\section{Results}
\label{sec:results}

\subsection{Cohort and feature completeness}

All 165 participants contributed complete values for AP-COP10,
Non-AP-COP25, Full35, and the three direct literature-derived COP
representations. No missing feature cells were present in these
representations. Within Full35, every candidate had at least 23 distinct
observed values, providing a basic check against degenerate or nearly
constant engineered features.

\subsection{Representation performance}
\label{subsec:performance}

The complete 35-candidate engineered representation (Full35) achieved
participant-level ROC AUC 0.908 (95\% CI 0.862--0.948), balanced accuracy
0.817, sensitivity 0.828, specificity 0.806, and Brier score 0.125. The
compact 10-candidate stance-indexed AP-COP representation (AP-COP10)
achieved AUC 0.894 (0.843--0.937), balanced accuracy 0.828, sensitivity
0.796, specificity 0.861, and Brier score 0.135
(Table~\ref{tab:main_performance}). Neither representation dominated across
all metrics: Full35 gave the highest AUC and the lowest Brier score,
whereas AP-COP10 gave higher fixed-threshold balanced accuracy and
specificity.

In the secondary nested representation analysis, adding the complementary
25 non-AP-COP descriptors to AP-COP10 increased AUC by 0.013, from 0.894
to 0.908 (95\% CI $-0.015$ to 0.043; $p=0.361$). Non-AP-COP25 evaluated
independently achieved AUC 0.856 (0.796--0.910), balanced accuracy 0.751,
sensitivity 0.753, specificity 0.750, and Brier score 0.154. Adding
AP-COP10 to Non-AP-COP25 increased AUC by 0.051 (95\% CI 0.015--0.089;
$p=0.0028$). AP-COP10 itself was numerically higher than Non-AP-COP25 by
0.038 AUC (95\% CI $-0.011$ to 0.090; $p=0.134$), although this direct
difference was not statistically supported. Thus, removing AP-COP10 from
Full35 produced a supported reduction in discrimination, whereas adding
the complementary descriptors to AP-COP10 produced only a small,
unsupported increase.

\begin{figure*}[t]
\centering
\includegraphics[width=0.95\textwidth]{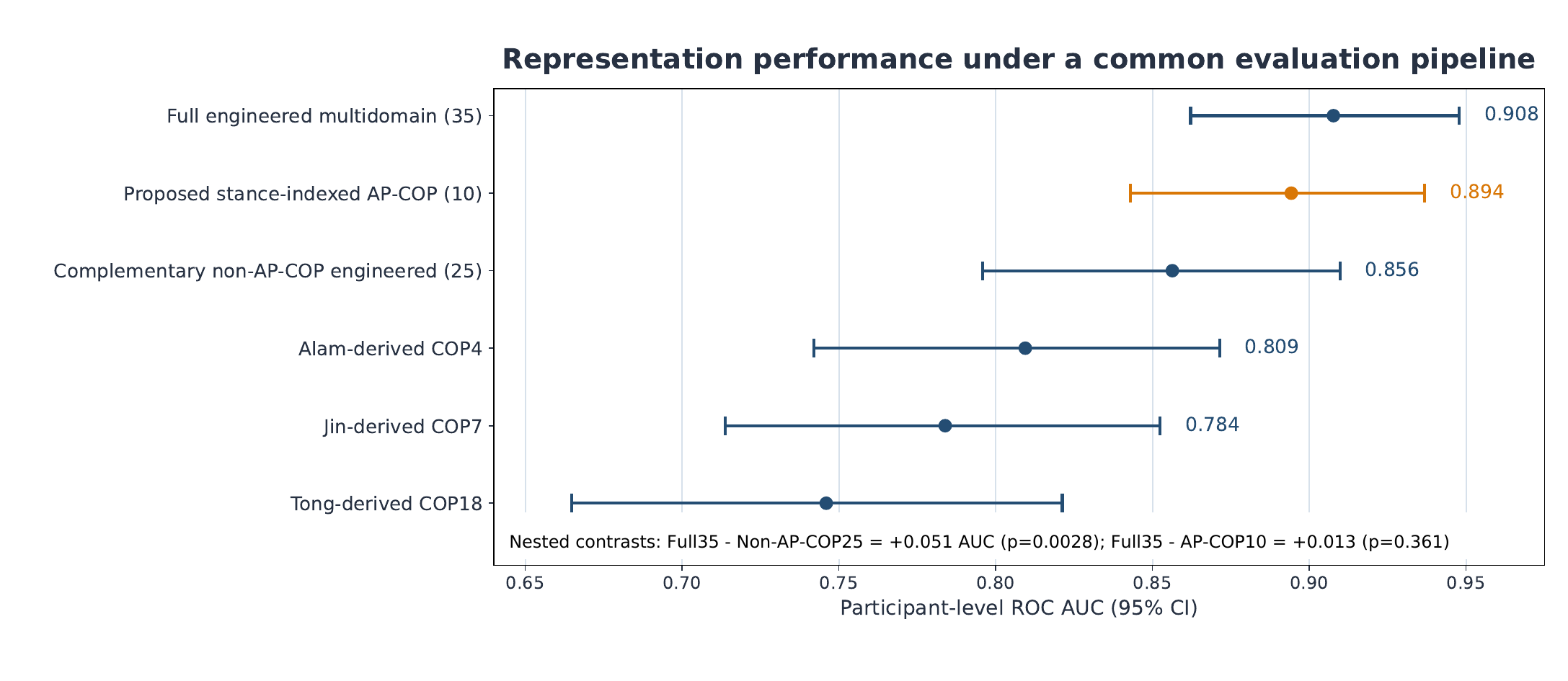}
\caption{Participant-level ROC AUC across the main engineered and
literature-derived representation spaces under the same fully nested
evaluation. Error bars show 95\% bootstrap confidence intervals.}
\label{fig:representation_performance}
\end{figure*}

Figure~\ref{fig:representation_performance} places these results alongside
the literature-derived comparators.

\begin{table*}[t]
\centering
\caption{Participant-level performance under repeated fully nested validation.}
\label{tab:main_performance}
\scriptsize
\setlength{\tabcolsep}{3.7pt}
\renewcommand{\arraystretch}{1.08}
\resizebox{0.99\textwidth}{!}{%
\begin{tabular}{lcccccccc}
\toprule
\textbf{Representation} &
\textbf{Candidates} &
\textbf{AUC (95\% CI)} &
\textbf{BAcc} &
\textbf{Sens.} &
\textbf{Spec.} &
\textbf{Brier} &
\textbf{$\Delta$AUC (AP-COP10 $-$ comparator) (95\% CI)} &
\textbf{$p$} \\
\midrule
Complete engineered multidomain (Full35)
& 35
& \textbf{0.908 (0.862--0.948)}
& 0.817
& \textbf{0.828}
& 0.806
& \textbf{0.125}
& $-0.013$ ($-0.043$ to 0.015)
& 0.361\textsuperscript{S} \\

\textbf{Proposed stance-indexed AP-COP (AP-COP10)}
& 10
& 0.894 (0.843--0.937)
& \textbf{0.828}
& 0.796
& \textbf{0.861}
& 0.135
& --
& -- \\

Complementary non-AP-COP engineered (Non-AP-COP25)
& 25
& 0.856 (0.796--0.910)
& 0.751
& 0.753
& 0.750
& 0.154
& 0.038 ($-0.011$ to 0.090)
& 0.134\textsuperscript{S} \\

Alam-derived COP4
& 4
& 0.809 (0.742--0.871)
& 0.705
& 0.688
& 0.722
& 0.180
& 0.085 (0.037--0.135)
& 0.0012\textsuperscript{H} \\

Jin-derived COP7
& 7
& 0.784 (0.714--0.852)
& 0.724
& 0.753
& 0.694
& 0.190
& 0.110 (0.050--0.171)
& 0.0012\textsuperscript{H} \\

Tong-derived COP18
& 18
& 0.746 (0.665--0.821)
& 0.700
& 0.677
& 0.722
& 0.206
& 0.148 (0.084--0.213)
& 0.0012\textsuperscript{H} \\
\bottomrule
\end{tabular}%
}

\vspace{2pt}
{\scriptsize
\textsuperscript{H}Holm-adjusted across the three direct
literature-derived COP comparisons.
\textsuperscript{S}Secondary nested representation comparison; not included
in the Holm-adjusted literature-comparison family.}
\end{table*}

\subsection{Controlled literature-grounded comparisons}

Under the identical pipeline, Alam-derived COP4 achieved AUC 0.809
(0.742--0.871), Jin-derived COP7 achieved 0.784 (0.714--0.852), and
Tong-derived COP18 achieved 0.746 (0.665--0.821). AP-COP10 achieved higher AUC, balanced accuracy, and specificity, and a
lower Brier score than all three literature-derived representations
(Table~\ref{tab:main_performance}).

The paired gain over Alam-derived COP4 was $\Delta$AUC $=0.085$ (95\% CI
0.037--0.135), over Jin-derived COP7 it was 0.110 (0.050--0.171), and over
Tong-derived COP18 it was 0.148 (0.084--0.213). Each raw bootstrap $p$
value was 0.0004; after Holm correction across the three direct literature
comparisons, all remained significant at
$p_{\mathrm{Holm}}=0.0012$.

Additional sensitivity analyses indicated that these gaps did not depend
on narrow encoding choices. A bilateral eight-candidate extension of the Alam
descriptor types achieved AUC 0.828 (0.762--0.889), with a paired
difference of 0.066 (0.025--0.110; $p=0.0024$). For the Tong family, the
bilateral-summary Tong9 representation achieved AUC 0.767, the AP-focused
Tong8 representation 0.758, and the axis-expanded Tong30 sensitivity
0.741; giving that descriptor family fewer, more AP-focused, or more
axis-explicit candidates did not approach 0.894. The 11-candidate
literature COP union achieved AUC 0.828 (0.764--0.887), 0.066 below
AP-COP10 (0.022--0.110; $p=0.0024$).

\subsection{Source-study holdout robustness}
\label{subsec:loso}

\begin{figure*}[t]
\centering
\includegraphics[width=0.95\textwidth]{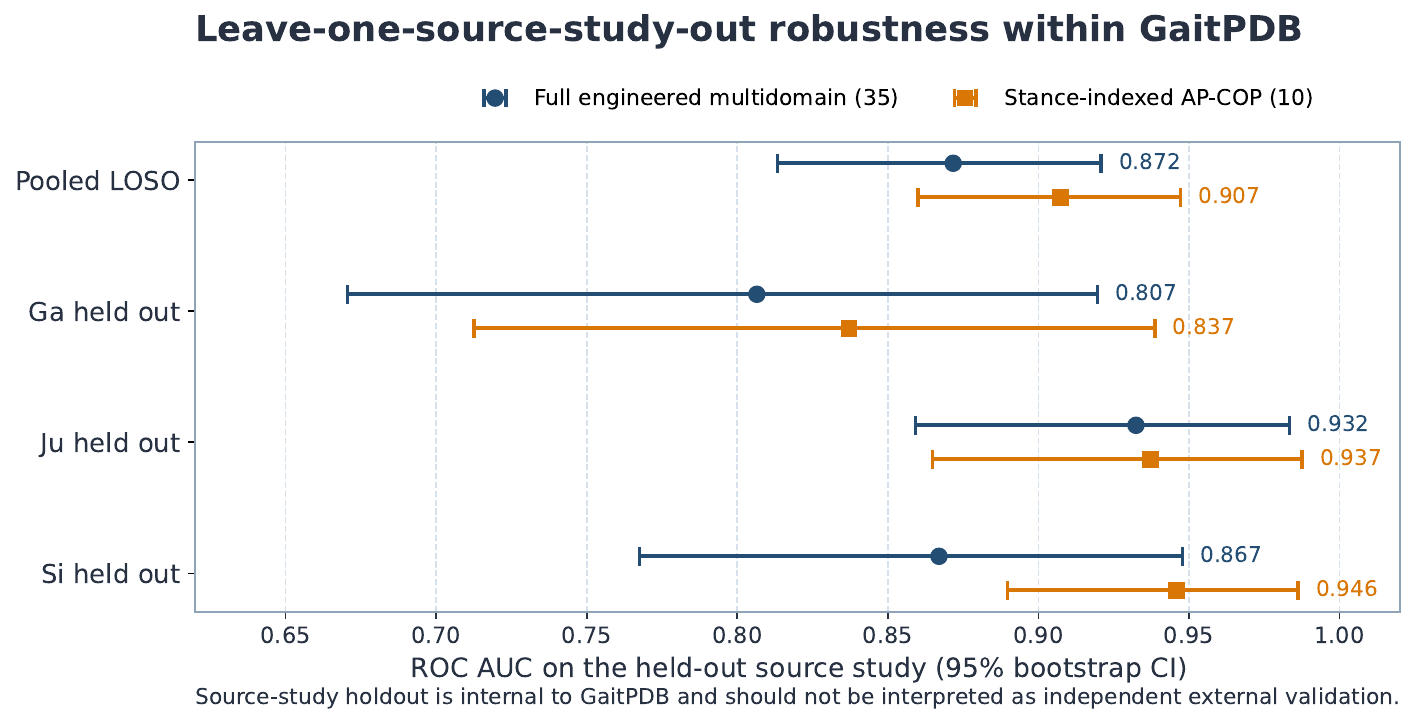}
\caption{Leave-one-source-study-out robustness within GaitPDB. ROC AUC and
95\% bootstrap confidence intervals are shown for Full35 and AP-COP10 when
each source study is excluded from model development.}
\label{fig:loso}
\end{figure*}

Leave-one-source-study-out evaluation removed each acquisition protocol
entirely from model development in turn. AP-COP10 achieved pooled AUC
0.907 (0.860--0.947), with 0.837 (0.713--0.939) when Ga was held out,
0.937 (0.865--0.988) for Ju, and 0.946 (0.890--0.986) for Si. Full35
achieved pooled AUC 0.872 (0.813--0.921), with 0.807 (0.670--0.920) for
Ga, 0.932 (0.859--0.983) for Ju, and 0.867 (0.767--0.948) for Si
(Figure~\ref{fig:loso}).

AP-COP10 had higher point estimates than Full35 in every held-out protocol
and in the pooled estimate. Its pooled holdout AUC (0.907) was slightly above its repeated
within-cohort estimate (0.894), whereas Full35 declined from 0.908 to
0.872 under the same procedure.

\subsection{External context: comparison with Jin's gait-only model}
\label{subsec:jin_context}

As external context for the source-study holdout results,
Table~\ref{tab:jin_loso_comparison} places the AP-COP10 per-protocol
holdout AUCs alongside the gait-only sensitivity model reported by Jin
\cite{jin2026}. AP-COP10 remained between 0.837 and 0.946 across the three
held-out protocols (mean 0.907), while Jin's gait-only model ranged from
0.500 to 0.845 (mean 0.726). The comparison is descriptive rather than
statistically paired: Jin's evaluation used a single 75/25 holdout split
rather than repeated nested cross-validation, and the gait-only feature set
retains non-COP demographic and contextual variables rather than being
restricted to COP descriptors alone.

\begin{table}[t]
\centering
\caption{Cross-protocol leave-one-source-study-out holdout AUC for the
proposed AP-COP10 representation and the gait-only sensitivity model
reported by Jin \cite{jin2026}.}
\label{tab:jin_loso_comparison}
\scriptsize
\setlength{\tabcolsep}{4pt}
\renewcommand{\arraystretch}{1.05}
\resizebox{\columnwidth}{!}{%
\begin{tabular}{lcccc}
\toprule
\textbf{Model} &
\textbf{Ga held out} &
\textbf{Ju held out} &
\textbf{Si held out} &
\textbf{Mean} \\
\midrule
Proposed AP-COP10
& 0.837 & 0.937 & 0.946 & \textbf{0.907} \\
Jin gait-only\textsuperscript{a} (SVM-RBF)
& 0.845 & 0.500 & 0.833 & 0.726 \\
\bottomrule
\end{tabular}
}

\vspace{2pt}
{\scriptsize
\textsuperscript{a}Retains non-clinical demographic/contextual variables
(age, weight, height, walking speed, trial index) alongside gait-derived
features; not restricted to COP descriptors.}
\end{table}

\subsection{Information across normalized stance}
\label{subsec:stance}

The strongest marginal PD--control differences occurred in early and early-mid stance.
AP-COP position at 25--35\% showed Hedges' $g=1.303$ and
direction-agnostic univariable AUC 0.828, while position at 5--15\%
showed $g=1.217$ and univariable AUC 0.810. Bilateral asymmetry at
5--15\% showed $g=0.842$. AP-COP position at 65--75\% showed no
statistically supported marginal separation ($g=-0.198$, 95\% CI
$-0.490$ to 0.095; $p=0.192$) yet was retained in 96\% of outer folds
within AP-COP10, and deleting it reduced AUC by 0.019
(95\% CI $-0.000$ to 0.041; $p=0.053$).

\begin{figure*}[t]
\centering
\includegraphics[width=0.94\textwidth]{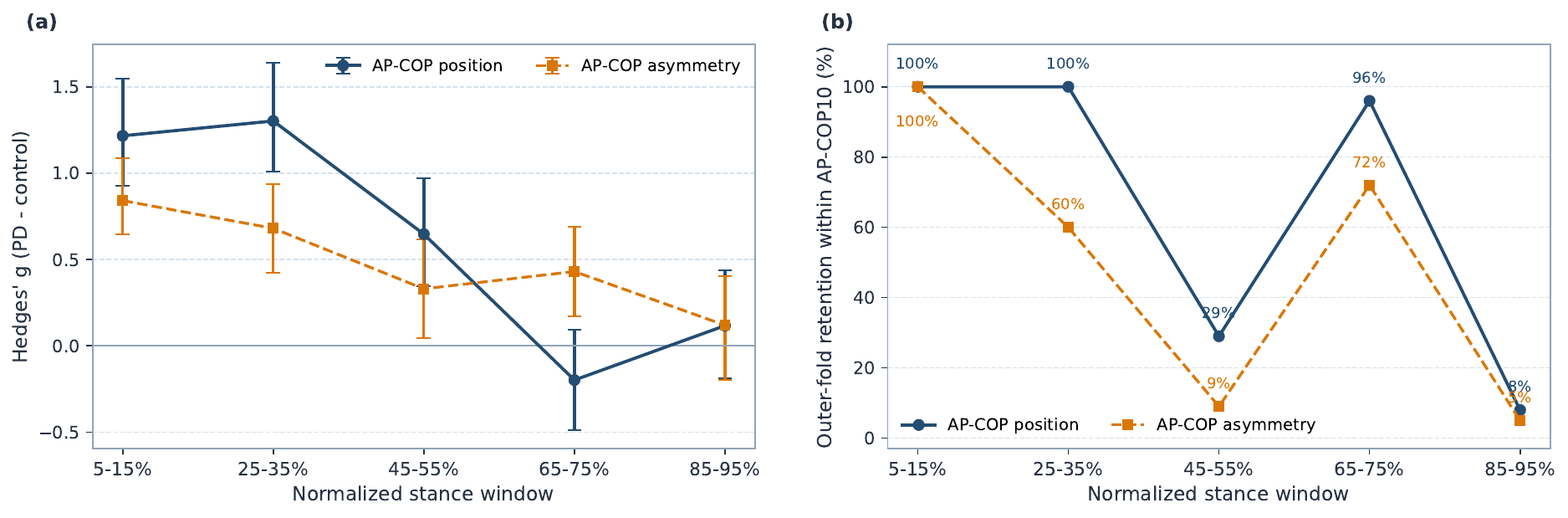}
\caption{AP-COP information across normalized stance. (a) Hedges' $g$ with
95\% confidence intervals for AP-COP position and bilateral asymmetry.
(b) Outer-fold retention frequencies within the stand-alone AP-COP10
representation.}
\label{fig:stance_results}
\end{figure*}

Within AP-COP10, outer-fold retention frequencies were 100\% for
5--15\% asymmetry, 100\% for 5--15\% position, 100\% for 25--35\%
position, 96\% for 65--75\% position, 72\% for 65--75\% asymmetry, and
60\% for 25--35\% asymmetry. Position at 45--55\% was retained in 29\%
of folds, while the remaining late-window candidates were selected less
frequently. Figure~\ref{fig:stance_results} summarizes marginal effects
and stand-alone retention.

\begin{figure*}[t]
\centering
\includegraphics[width=0.95\textwidth]{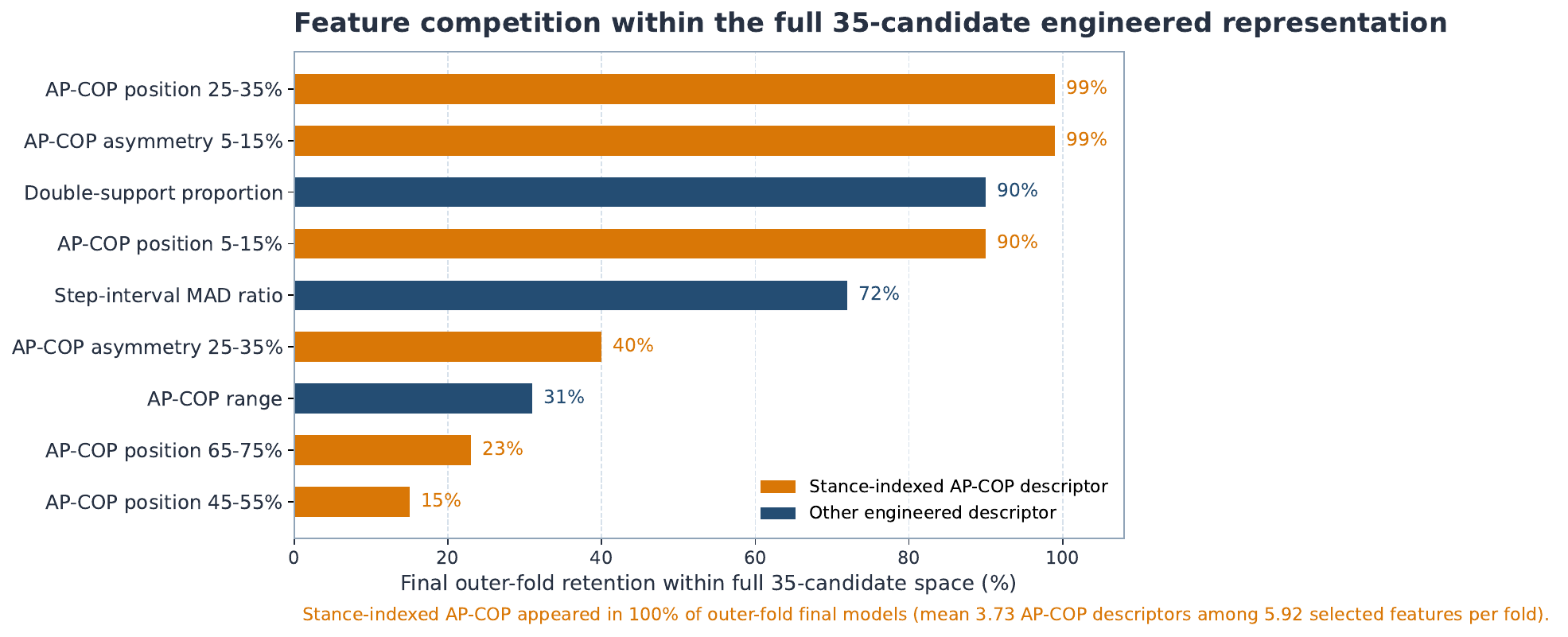}
\caption{Feature competition within the complete 35-candidate engineered
representation. Bars show final outer-fold retention after training-only
stability selection, redundancy pruning, and the six-feature cap.
Stance-indexed AP-COP descriptors are highlighted.}
\label{fig:competition}
\end{figure*}

The literature-derived selections provide a face-validity comparison. In
Alam-derived COP4, AP-COP SD and AP-COP mean were retained in every outer
fold, while the ML quantities were less stable. In Jin-derived COP7, left
AP-COP SD was retained in every fold and right AP-COP SD in 98\%, while
the remaining path and ML descriptors were rarely retained. In
Tong-derived COP18, AP-coordinate and AP-entropy/velocity quantities were
among the most stable. Across distinct representation families, the AP
direction therefore repeatedly emerged as informative; AP-COP10 differs
by preserving AP-COP \emph{location across stance} rather than compressing
it primarily into whole-trajectory summaries.

\begin{table}[t]
\centering
\caption{Outer-fold retention of the 10 stance-indexed AP-COP descriptors
under competition within Full35, with the most frequently retained
descriptors from other families for context.}
\label{tab:competition}
\scriptsize
\setlength{\tabcolsep}{3.4pt}
\renewcommand{\arraystretch}{1.03}
\begin{tabular}{lcc}
\toprule
\textbf{Descriptor} & \textbf{Family} & \textbf{Retention} \\
\midrule
Asymmetry 5--15\%        & AP-COP       & 0.99 \\
Position 25--35\%        & AP-COP       & 0.99 \\
Position 5--15\%         & AP-COP       & 0.90 \\
Asymmetry 25--35\%       & AP-COP       & 0.40 \\
Position 65--75\%        & AP-COP       & 0.23 \\
Position 45--55\%        & AP-COP       & 0.15 \\
Asymmetry 65--75\%       & AP-COP       & 0.05 \\
Position 85--95\%        & AP-COP       & 0.01 \\
Asymmetry 45--55\%       & AP-COP       & 0.01 \\
Asymmetry 85--95\%       & AP-COP       & 0.00 \\
\midrule
Double-support proportion    & Support      & 0.90 \\
Step-interval irregularity   & Temporal     & 0.72 \\
COP AP range                 & COP rollover & 0.31 \\
COP path length              & COP rollover & 0.05 \\
Median COP speed             & COP rollover & 0.03 \\
\bottomrule
\end{tabular}
\end{table}

\subsection{Feature competition within the complete representation}
\label{subsec:competition}

Placing all 10 stance-indexed descriptors in the same candidate pool as
25 descriptors from five other biomechanical families allows direct
competition under training-only selection. Final models retained a mean
of 5.92 descriptors per outer fold (median 6), with mean pairwise Jaccard
similarity of 0.62 between fold-specific selections (median 0.71).

The stance-indexed AP-COP family was represented in every outer-fold final
model, contributing a mean of 3.73 of the 5.92 selected descriptors
(Table~\ref{tab:competition}). That contribution was highly concentrated
rather than distributed across the family. Three descriptors dominated:
bilateral asymmetry at 5--15\% (99\% retention), position at 25--35\%
(99\%), and position at 5--15\% (90\%). Asymmetry at 25--35\% was
retained in 40\% of folds and position at 65--75\% in 23\%, while the
remaining five stance-indexed descriptors were retained in at most 15\%
of folds. Among descriptors outside the AP-COP family, double-support
proportion was retained in 90\% of folds and step-interval irregularity
in 72\%; the six broader COP-rollover descriptors were retained in at
most 31\% of folds (AP range), with COP path length at 5\% and median
COP speed at 3\%.

The retention profile for position at 65--75\% differs markedly between
the two settings: 96\% within the stand-alone AP-COP10 representation
and 23\% under Full35 competition, displaced once double-support and
temporal-rhythm descriptors enter the same pool
(Figure~\ref{fig:competition}).

\subsection{Attribution and speed sensitivity}

Secondary attribution analyses examined the relative contribution of
position and asymmetry descriptors within AP-COP10. A five-position-only
representation achieved AUC 0.857, whereas the five asymmetry descriptors
alone achieved 0.746. Adding only the early 5--15\% asymmetry descriptor
to the five position features increased AUC to 0.900. Attribution
therefore points to broad stance-indexed position coverage as the primary
source of discriminative information, with early-stance asymmetry
contributing complementary value beyond position alone. This pattern is
consistent with the retention profile observed under Full35 competition,
where the three most frequently retained stance-indexed descriptors were
early-stance asymmetry and the 25--35\% and 5--15\% positions.

Walking speed alone achieved AUC 0.772, and age with speed 0.767. Adding
speed to AP-COP10 increased AUC to 0.919, and adding both age and speed
yielded 0.922.

The relative asymmetry index is coordinate-origin dependent. Replacing it
with the absolute bilateral AP-COP difference produced AUC 0.896, and the
AP-range-normalized bilateral difference produced 0.894.

\subsection{Compactness, classifier family, and sensor sensitivity}

Within Full35, AUC rose from 0.802 with a one-descriptor maximum to 0.865
with two, 0.885 with three, 0.901 with four, 0.906 with five, and 0.908
with six. The paired gain of the six-descriptor budget over three
descriptors was 0.022 (0.002--0.045; $p=0.033$), whereas its gains over
four and five descriptors were not statistically supported
(0.006, $p=0.264$; 0.001, $p=0.699$). Rerunning fold-specific selection
with maximum final-feature caps of 4, 6, 8, 10, and 12 produced AUCs of
0.901, 0.908, 0.908, 0.905, and 0.903; none of the larger caps improved
on six (cap 6 minus cap 8: $-0.001$, $p=0.864$; minus cap 10:
0.002, $p=0.543$; minus cap 12: 0.005, $p=0.211$)
(Supplementary Table~S8).

Classifier-family sensitivity on the same fold-specific descriptors
produced AUC 0.913 for histogram gradient boosting, 0.908 for logistic
regression, 0.899 for RBF-SVM, and 0.898 for random forest. Calibration
slopes were 1.01 for logistic regression and 0.73 for histogram gradient
boosting (Supplementary Table~S9).

Simulated sensor-gain variation produced graded degradation: mean AUC
fell from 0.908 on clean input to 0.907 at 5\% gain variation, 0.902 at
10\%, and 0.881 at 20\%. Random single-sensor dropout reduced mean AUC
to 0.858 and two-sensor dropout to 0.802. Under 10\% gain variation,
the five position descriptors retained median Spearman agreement ranging
from 0.985 to 0.996 with their clean values, whereas the five asymmetry
descriptors ranged from 0.865 to 0.921. Under single-sensor dropout,
position descriptors ranged from 0.831 to 0.966, while asymmetry
descriptors ranged from 0.544 to 0.618
(Supplementary Figure~S1; Supplementary Table~S10).

\section{Discussion}
\label{sec:discussion}

\subsection{Principal finding: representation matters}

The central result is a controlled representation effect. When the
classifier, outer splits, feature-selection procedure, and participant
cohort were held constant, AP-COP10 provided higher discrimination than
three distinct COP descriptor families derived from prior GaitPDB studies.
The gain was 0.085 AUC over Alam-derived global coordinate statistics,
0.110 over Jin-derived dispersion/path/symmetry descriptors, and 0.148
over the primary Tong-derived dynamic/nonlinear descriptor set; all three
differences remained significant after Holm correction. The result does
not rely on comparing our cross-validation AUC with the published headline
accuracy of another paper. It instead addresses a narrower and more
defensible question: how much predictive information is retained by
different COP representations when evaluated under the same
machine-learning procedure.

This distinction matters because the source studies address different
scientific questions. Alam et al. combined COP with temporal and force
features \cite{alam2017}. Jin's full model included clinical severity and
demographic variables, and even the gait-only sensitivity included non-COP
quantities such as walking speed \cite{jin2026}. Tong et al. addressed
multiclass severity classification with topological processing and
oversampling \cite{tong2021}. Their reported performance therefore cannot
be ranked directly against the present binary participant-level
experiment. The harmonized reimplementation isolates the COP
representation itself.

\subsection{Compactness within a complete engineered feature space}

The nested representation comparisons strengthen the compactness result.
Full35 achieved AUC 0.908, AP-COP10 achieved 0.894, and Non-AP-COP25
achieved 0.856. Adding the 25 non-AP-COP descriptors to AP-COP10 increased
AUC by only 0.013 (95\% CI $-0.015$ to 0.043; $p=0.361$). In the reverse
comparison, adding AP-COP10 to Non-AP-COP25 increased AUC by 0.051
(95\% CI 0.015--0.089; $p=0.0028$). AP-COP10 was also numerically higher
than Non-AP-COP25 by 0.038 AUC, although that direct difference was not
statistically supported ($p=0.134$). The results therefore do not indicate
that the complementary 25 descriptors contain no predictive information;
their stand-alone AUC of 0.856 clearly shows that they do. Rather, they
suggest substantial overlap between the information captured by AP-COP10
and the broader engineered descriptors, while the supported loss after
removing AP-COP10 indicates that part of the stance-indexed information is
not readily recovered by the complementary feature bank. No equivalence
margin was prespecified, so the Full35--AP-COP10 null contrast remains a
null result rather than evidence of equivalence.

The feature-budget, selection, and cap analyses reinforce this
interpretation. Non-AP-COP25 selected a mean of 5.91 descriptors per outer
fold (median 6), indicating that its lower AUC was not caused by
systematically sparse final models. Within Full35, stance-indexed AP-COP
appeared in every final model and contributed a mean of 3.73 of the 5.92
retained descriptors. Four to six retained descriptors recovered nearly
all available Full35 discrimination, and allowing 8, 10, or 12 final
descriptors produced no supported improvement over six. Together, these
results indicate that substantial discriminative information in the broad
engineered gait bank can be represented using a small number of
interpretable descriptors, with stance-indexed AP-COP accounting for a
large share of the selected features.

The classifier-family analysis points in the same direction. Histogram
gradient boosting produced only a small numerical AUC increase over
logistic regression (0.913 versus 0.908), while RBF-SVM and random forest
achieved 0.899 and 0.898, respectively. Calibration favored the linear
reference, with a slope of 1.01 for logistic regression compared with 0.73
for histogram gradient boosting. This should not be interpreted as
evidence against deep or graph-based gait models, which have demonstrated
strong VGRF discrimination \cite{elmaachi2020,choi2026,wang2026graph}.
Rather, for this controlled representation-comparison task, careful
feature construction concentrated substantial predictive information
without requiring a highly flexible final classifier.

\subsection{Which stance locations survive competition}

Under nested competition within Full35, the stance-indexed family appeared
in every final model, but its contribution was concentrated in three
descriptors: early-stance asymmetry, position at 25--35\%, and position at
5--15\%. Position at 25--35\% was retained in 99\% of outer folds,
ranking above every temporal, support, loading, and broader-COP descriptor
in the candidate pool. This provides strong evidence that stance-localized
AP-COP position retains information that is not fully displaced by the
other engineered descriptors with which it competes. The complete
outer-fold retention profiles are provided in Supplementary
Tables~S4--S5.

The remaining seven stance-indexed descriptors were selected less often
under full competition, and one case is particularly informative.
Position at 65--75\% was retained in 96\% of folds within AP-COP10 but
only 23\% within Full35. Removing it from AP-COP10 reduced AUC by 0.019
(95\% CI $-0.000$ to 0.041; $p=0.053$). This descriptor also showed no
statistically supported univariable separation on its own
(Hedges' $g=-0.198$), illustrating that marginal effect size and
multivariable usefulness are not necessarily equivalent. A plausible
interpretation is that the descriptor contributes conditional information
about later stance when only AP-COP descriptors are available, whereas
other temporal or support descriptors may capture overlapping information
once they enter the broader candidate pool. This also cautions against
interpreting stand-alone retention frequency as evidence of intrinsic
descriptor importance.

The broader COP-rollover descriptors behaved differently from the
localized positions. AP range was retained in 31\% of folds, whereas path
length, median speed, speed variability, monotonicity, and early slope were
retained much less frequently under Full35 competition. This pattern is
consistent with the controlled literature comparisons, although
correlation and competition among candidates mean that retention
frequencies should not be interpreted as independent evidence that
whole-trajectory summaries are inherently less informative.

\subsection{From global AP-COP summaries to stance-indexed position}

The literature-derived results show substantial agreement rather than
contradiction. Across the Alam-, Jin-, and Tong-derived feature sets,
AP-direction descriptors were repeatedly more consistently informative than many ML or
path-based alternatives. Jin identified reduced AP-COP dispersion as one
of the strongest biomechanical effects in GaitPDB \cite{jin2026}, while
Tong reported that COP quantities in the walking direction were important
for severity differentiation \cite{tong2021}. The present results support
the same broad biomechanical direction but extend it representationally:
instead of summarizing only how much AP-COP varies, how long its path is,
or how irregular its derivatives are, AP-COP10 preserves \emph{where} the
AP load center is at multiple fixed locations across normalized stance.

Independent waveform evidence from Jin also supports the broader premise
that plantar-force differences can be localized rather than uniform across
stance. Jin's statistical-parametric analysis of VGRF---a different signal
from COP---identified significant PD--control clusters in early stance
(approximately 0--13\% on the left and 6--12\% on the right), midstance
on the right (approximately 34--55\%), and late stance
(approximately 90--96\%) \cite{jin2026}. These intervals should not be
interpreted as validating the present COP windows, but they independently
reinforce the rationale for retaining stance location rather than relying
only on whole-trajectory scalars.

The present study does not claim that examining COP across stance is
itself novel. COP progression has long been analyzed as a trajectory,
across stance conditions, and within stance subregions
\cite{han1999,lugade2014,chiu2013speed,buldt2018,pataky2014}. Recent
plantar-pressure studies have likewise explored regional and multidomain
representations for PD \cite{sun2023,nanayakkara2025}. The contribution
here is the compact five-window AP-COP encoding, its controlled comparison
with multiple GaitPDB-derived COP summary families, and its behavior when
placed in direct competition with a broader engineered feature space.

\subsection{Source-protocol robustness}

As shown in Section~\ref{subsec:jin_context}, Jin's gait-only sensitivity
model \cite{jin2026} reached AUC 0.500 when the Ju protocol was held out,
despite retaining non-COP demographic and contextual variables alongside
gait-derived features. In the present study, AP-COP10 remained between
0.837 and 0.946 across the three held-out source studies, with pooled AUC
0.907, and its point estimate was higher than Full35 in every holdout. Full35
declined from AUC 0.908 under repeated within-resource validation to 0.872
under source-study holdout, whereas AP-COP10 changed from 0.894 to a pooled
holdout AUC of 0.907.

One possible interpretation is that a larger candidate pool provides more
opportunity to select descriptors whose behavior is associated with
source-study characteristics, whereas restricting the representation to
one signal family may limit that exposure. The present results do not
establish this mechanism, however, and the confidence intervals overlap.
Moreover, Ga, Ju, and Si are all contained within GaitPDB and share the
same general sensing platform and sensor geometry. Leave-one-source-study-
out evaluation therefore measures robustness to acquisition protocol
within one database and should not be interpreted as external validation.
Independent cohorts and different sensing layouts remain necessary.

\subsection{Asymmetry, walking speed, and measurement robustness}

Bilateral asymmetry is biologically plausible in PD because motor
manifestations are often lateralized, and postural and gait asymmetry have
been reported previously \cite{geurts2011,beretta2015,shin2020}. In
AP-COP10, asymmetry is aligned to the same stance windows as position.
Early-stance asymmetry was retained in 99\% of Full35 outer folds and
provided complementary information when added to the five position
features, whereas the five asymmetry descriptors alone were substantially
weaker than the position-only representation. The relative asymmetry index
also depends on coordinate origin. Origin-invariant variants produced
nearly identical AUCs, supporting the overall representation conclusion
without implying that the relative asymmetry index itself is an
origin-invariant biomechanical measure.

Walking speed is an important interpretive issue because COP progression
changes with speed \cite{chiu2013speed}, and PD gait rhythmicity is also
speed dependent \cite{frenkel2005}. Speed alone achieved AUC 0.772,
whereas AP-COP10 combined with speed achieved 0.919. These results indicate
that AP-COP10 contains predictive information beyond that provided by
walking speed alone. They do not establish causal independence from speed,
age, disease severity, or other participant characteristics, and the
representation should not be interpreted as a pure disease mechanism.

The sensor perturbation analysis provides a separate measurement
perspective. Under single-sensor dropout, AP-COP position descriptors were
more stable overall, with median Spearman agreement ranging from 0.831 to
0.966, than the asymmetry descriptors, which ranged from 0.544 to 0.618.
The terminal-stance position descriptor was more sensitive than the other
position features. Early-stance asymmetry remained important predictively,
being retained in 99\% of Full35 outer folds, but its dropout stability was
only $\rho=0.618$. The lowest asymmetry stability occurred at 65--75\%
stance ($\rho=0.544$). Predictive usefulness and measurement robustness
are therefore distinct properties, particularly for bilateral asymmetry
features. These simulations held contact detection and stance segmentation
fixed and used five random realizations per condition; they should be
interpreted as a first-order robustness check rather than device-failure
testing.

\subsection{Limitations}
\label{subsec:limitations}

Several limitations define the scope of the findings. First, all
participants were drawn from one historical public database and one
plantar-sensor platform. Although GaitPDB contains three source studies,
leave-one-source-study-out evaluation remains internal to the same
resource and does not constitute external validation. Only one
usual-walking recording was retained per participant, so test--retest
reliability was not assessed. Independent cohorts, repeated measurements,
and different sensing platforms are therefore needed to establish broader
generalizability \cite{mancini2025,wang2026survey}.

Second, participant characteristics may contribute to the observed
discrimination. Walking speed alone was predictive, and the present
speed and age analyses assess predictive information rather than causal
covariate adjustment. Effects of walking speed, age, disease severity,
medication state, comorbidities, and other cohort characteristics therefore
cannot be excluded. The reported representation should consequently not
be interpreted as isolating a PD-specific biomechanical mechanism.

Third, AP-COP was reconstructed as a sensor-coordinate force-weighted
load-center proxy tied to the GaitPDB insole geometry rather than as an
anatomical COP distance from a continuous pressure field. The five stance
windows were fixed before the reported classification analyses, but their
optimal locations and widths have not been established. Transfer to other
sensor geometries or alternative stance-window definitions may therefore
require additional validation.

Finally, the literature-derived comparators are harmonized
reimplementations of reproducible COP descriptor definitions rather than
exact replications of the complete published pipelines, and Full35 and
Non-AP-COP25 represent finite engineered feature banks rather than all
possible gait representations. The supported loss after removing AP-COP10
therefore applies to the engineered descriptors tested here and does not
exclude additional information in richer waveform-based or learned
representations. Sensor perturbations were also simulated rather than
tested on physically degraded hardware. More broadly, the task is
established PD versus healthy control classification and should be
interpreted as an evaluation of representation quality rather than as
validation of a clinical diagnostic system.

\section{Conclusion}
\label{sec:conclusion}

This study demonstrates that how plantar-force signals are represented can
substantially influence the information available for machine-learning
analysis of Parkinsonian gait. Preserving anterior--posterior COP position
at corresponding locations across normalized stance produced a compact
representation that consistently performed strongly against conventional
COP summaries and remained informative when placed in competition with a
broader engineered gait feature space. The complementary representation
analyses further indicate that the information carried by stance-indexed
AP-COP is not readily recovered when these descriptors are removed, even
though the remaining temporal, support, asymmetry, COP-rollover, and
loading features retain substantial discriminative information.

The findings therefore support a shift from treating COP primarily as a
trajectory to be summarized globally toward preserving \emph{where} its
informative behavior occurs during stance. In this representation, AP-COP
position provides the main discriminative structure, with bilateral
asymmetry contributing complementary information, particularly in early
stance. Overall, AP-COP10 provides a compact and interpretable way to
retain localized gait information without relying on a large engineered
feature bank. Validation in independent cohorts and different sensing
platforms is the necessary next step before broader clinical or
translational conclusions can be drawn.

\section*{Data and code availability}
The GaitPDB dataset is publicly available through PhysioNet \cite{hausdorff2008}. The analysis notebooks and derived result archives used for the complete engineered representation can be made available from the authors upon reasonable request. The reported analyses were executed with Python 3.10.19, NumPy 1.26.4, and pandas 2.3.3 on Linux. Participant-level predictions, feature-selection summaries, paired comparisons, and supporting numerical outputs are retained by the authors.

\section*{Ethics statement}
This study is a secondary analysis of deidentified, publicly available data. No new participants were recruited and no intervention was performed. Ethics approval and informed-consent procedures for the original data collections are described in the source-study publications and database documentation.

\section*{CRediT authorship contribution statement}

The authors' contributions are described according to the Contributor Roles Taxonomy (CRediT):

\noindent\textbf{Md. Sifat:}
Conceptualization, Methodology, Software, Formal analysis, Investigation,
Writing -- original draft, Writing -- review \& editing.

\noindent\textbf{Sania Akter:}
Data curation, Investigation, Validation, Writing -- review \& editing.

\noindent\textbf{Akif Islam:}
Methodology, Validation, Resources, Writing -- review \& editing.

\noindent\textbf{Md. Ekramul Hamid:}
Supervision, Project administration, Writing -- review \& editing.

All authors have read and approved the final version of the manuscript and
agree to be accountable for all aspects of the work.

\section*{Supplementary Material}
\setcounter{section}{0}
\setcounter{subsection}{0}
\setcounter{table}{0}
\setcounter{figure}{0}
\renewcommand{\thesection}{S\arabic{section}}
\renewcommand{\thesubsection}{\thesection.\arabic{subsection}}
\renewcommand{\thetable}{S\arabic{table}}
\renewcommand{\thefigure}{S\arabic{figure}}

\section{Scope and reproducibility}

This supplement has two purposes. First, it records the exact candidate-feature identifiers used in the final analyses and documents the literature-derived COP reference sets. Second, it provides supporting analyses that probe compactness, selection stability, classifier dependence, sensor perturbation, asymmetry normalization, covariate sensitivity, and source-study robustness. The Holm-adjusted literature-comparison family comprises AP-COP10 versus the three literature-derived COP representations. The AP-COP10--Full35 and Non-AP-COP25 contrasts are secondary nested representation analyses and were not included in that Holm-adjusted family. All supporting model analyses retain the same separation between model development and outer-fold performance estimation used in the main manuscript \cite{varma2006,cawley2010}.

\section{Feature dictionary}

\subsection{Proposed stance-indexed AP-COP10}

The proposed representation contains five bilateral AP-COP position descriptors and five bilateral asymmetry descriptors at fixed normalized-stance windows

\[
W=\{5\text{--}15,\;25\text{--}35,\;45\text{--}55,\;65\text{--}75,\;85\text{--}95\}\%.
\]

The identifiers below are the exact feature-column names used in the final analysis.

\begin{table}[H]

\centering

\caption{Exact feature identifiers for AP-COP10.}\label{tab:apcop10}

\scriptsize

\setlength{\tabcolsep}{3pt}

\begin{tabularx}{\columnwidth}{p{0.50cm}p{2.92cm}p{1.7cm}X}

\toprule

ID & Feature identifier & Window & Role\\

\midrule

AP1 & \texttt{cop\_landing\_position} & 5--15\% & Bilateral AP-COP position\\

AP2 & \texttt{cop\_pos\_25\_35} & 25--35\% & Bilateral AP-COP position\\

AP3 & \texttt{cop\_midstance\_position} & 45--55\% & Bilateral AP-COP position\\

AP4 & \texttt{cop\_pos\_65\_75} & 65--75\% & Bilateral AP-COP position\\

AP5 & \texttt{cop\_terminal\_position} & 85--95\% & Bilateral AP-COP position\\

AP6 & \texttt{cop\_landing\_asymmetry} & 5--15\% & Bilateral AP-COP asymmetry\\

AP7 & \texttt{cop\_asym\_25\_35} & 25--35\% & Bilateral AP-COP asymmetry\\

AP8 & \texttt{cop\_asym\_45\_55} & 45--55\% & Bilateral AP-COP asymmetry\\

AP9 & \texttt{cop\_asym\_65\_75} & 65--75\% & Bilateral AP-COP asymmetry\\

AP10 & \texttt{cop\_asym\_85\_95} & 85--95\% & Bilateral AP-COP asymmetry\\

\bottomrule

\end{tabularx}

\end{table}

For each window, stance-level AP-COP is averaged within the window; repeated stances are summarized separately for the two feet by their median; bilateral position is the mean of the two side-specific summaries; and bilateral asymmetry is computed using the primary relative index defined in the main manuscript.

\clearpage
\onecolumn

\begin{landscape}

\begin{table}[p]
\centering
\caption{Complete Full35 feature dictionary grouped by descriptor family.}
\label{tab:full35_dictionary}

\scriptsize
\setlength{\tabcolsep}{5pt}
\renewcommand{\arraystretch}{1.05}

\begin{tabularx}{0.96\linewidth}{
    >{\raggedright\arraybackslash}p{0.7cm}
    >{\ttfamily\raggedright\arraybackslash}p{5.4cm}
    >{\raggedright\arraybackslash}p{3.5cm}
    X
}
\toprule
\textbf{ID} &
\textbf{Feature identifier} &
\textbf{Family} &
\textbf{Information represented} \\
\midrule

\multicolumn{4}{l}{\textit{Temporal/rhythm}} \\
1 & step\_interval\_median
  & Temporal/rhythm
  & Median alternating-foot step interval. \\

2 & step\_interval\_cv
  & Temporal/rhythm
  & Step-interval variability. \\

3 & step\_interval\_mad\_ratio
  & Temporal/rhythm
  & Step-interval MAD relative to central scale. \\

4 & stride\_interval\_cv
  & Temporal/rhythm
  & Same-foot stride-interval variability. \\

5 & stance\_duration\_cv
  & Temporal/rhythm
  & Stance-duration variability. \\

6 & swing\_duration\_cv
  & Temporal/rhythm
  & Swing-duration variability. \\

\addlinespace[1pt]

\multicolumn{4}{l}{\textit{Dynamic support}} \\
7 & double\_support\_proportion
  & Dynamic support
  & Proportion of the gait cycle in double support. \\

8 & double\_support\_duration\_median
  & Dynamic support
  & Median double-support duration. \\

9 & stance\_ratio\_median
  & Dynamic support
  & Median stance-ratio descriptor. \\

10 & stance\_duration\_median
   & Dynamic support
   & Median stance duration. \\

\addlinespace[1pt]

\multicolumn{4}{l}{\textit{Bilateral asymmetry outside AP-COP10}} \\
11 & stance\_ratio\_asymmetry
   & Bilateral asymmetry
   & Left--right stance-ratio asymmetry. \\

12 & stride\_interval\_asymmetry
   & Bilateral asymmetry
   & Left--right stride-interval asymmetry. \\

13 & stance\_duration\_asymmetry
   & Bilateral asymmetry
   & Left--right stance-duration asymmetry. \\

14 & peak\_force\_asymmetry
   & Bilateral asymmetry
   & Left--right peak-force asymmetry. \\

\addlinespace[1pt]

\multicolumn{4}{l}{\textit{Stance-indexed AP-COP10}} \\
15 & cop\_landing\_position
   & AP-COP position
   & AP-COP position at 5--15\%. \\

16 & cop\_pos\_25\_35
   & AP-COP position
   & AP-COP position at 25--35\%. \\

17 & cop\_midstance\_position
   & AP-COP position
   & AP-COP position at 45--55\%. \\

18 & cop\_pos\_65\_75
   & AP-COP position
   & AP-COP position at 65--75\%. \\

19 & cop\_terminal\_position
   & AP-COP position
   & AP-COP position at 85--95\%. \\

20 & cop\_landing\_asymmetry
   & AP-COP asymmetry
   & AP-COP asymmetry at 5--15\%. \\

21 & cop\_asym\_25\_35
   & AP-COP asymmetry
   & AP-COP asymmetry at 25--35\%. \\

22 & cop\_asym\_45\_55
   & AP-COP asymmetry
   & AP-COP asymmetry at 45--55\%. \\

23 & cop\_asym\_65\_75
   & AP-COP asymmetry
   & AP-COP asymmetry at 65--75\%. \\

24 & cop\_asym\_85\_95
   & AP-COP asymmetry
   & AP-COP asymmetry at 85--95\%. \\

\addlinespace[1pt]

\multicolumn{4}{l}{\textit{Broader COP rollover}} \\
25 & cop\_ap\_range
   & COP rollover
   & AP-COP range across stance. \\

26 & cop\_path\_length
   & COP rollover
   & Two-dimensional COP path length. \\

27 & cop\_speed\_median
   & COP rollover
   & Median COP speed. \\

28 & cop\_speed\_cv
   & COP rollover
   & COP-speed variability. \\

29 & cop\_monotonicity
   & COP rollover
   & Monotonicity of COP progression. \\

30 & cop\_early\_slope
   & COP rollover
   & Early-stance COP slope. \\

\addlinespace[1pt]

\multicolumn{4}{l}{\textit{Vertical loading}} \\
31 & peak\_force\_normalized
   & Vertical loading
   & Normalized peak plantar force. \\

32 & force\_impulse\_normalized
   & Vertical loading
   & Normalized force impulse. \\

33 & loading\_rate\_normalized
   & Vertical loading
   & Normalized loading rate. \\

34 & unloading\_rate\_normalized
   & Vertical loading
   & Normalized unloading rate. \\

35 & force\_shape\_variability
   & Vertical loading
   & Variability of normalized force-shape descriptor. \\

\bottomrule
\end{tabularx}

\end{table}

\end{landscape}

\begin{table*}[!t]
\centering
\caption{Literature-derived COP reproduction record.}
\label{tab:lit_reproduction}
\small
\setlength{\tabcolsep}{5pt}
\begin{tabularx}{0.98\textwidth}{p{1.2cm}p{0.6cm}X}
\toprule

  Set & $n$ & Reproduced descriptors and harmonization decisions\\

  \midrule

  Alam4 & 4 & Mean/SD ML-COP and mean/SD AP-COP from the left-foot reference. A bilateral eight-candidate extension was additionally evaluated as a sensitivity analysis.\\

  Jin7 & 7 & Left/right AP-COP SD, left/right ML-COP SD, left/right 2-D COP path length, and bilateral COP-path symmetry. Forefoot-to-rearfoot loading was excluded because the controlled comparison was restricted to COP-specific descriptors; Jin7 is therefore not the complete source-paper model.\\

  Tong18 & 18 & Nine descriptor types per foot: RMS ML-COP, RMS AP-COP, RMS total COP magnitude, RMS 2-D COP velocity, RMS acceleration, RMS jerk, COP path efficiency, ML-COP sample entropy, and AP-COP sample entropy. CSIP/CISP were excluded because their reconstruction was under-specified; source-specific persistent-homology processing, oversampling, and SVM classification were not reproduced. Sample entropy used $m=2$ and $r=0.20\times$SD.\\

  \bottomrule
\end{tabularx}
\end{table*}

\renewcommand{\arraystretch}{0.94}

\section{Literature-derived COP feature reproduction}

\label{sec:lit_reproduction}

The literature-derived sets are harmonized reimplementations of published COP descriptor definitions, not reproductions of the original end-to-end classifiers. The common participant-level aggregation, preprocessing, nested feature selection, and final classifier are those of the main manuscript.

An 11-candidate Alam--Jin union was also assembled as a secondary comparison and was not included in the Holm-adjusted literature-comparison family. Additional encoding sensitivities were evaluated for the Alam and Tong descriptor families. The bilateral Alam8 extension achieved AUC 0.828 (95\% CI 0.762--0.889), with AP-COP10 higher by 0.066 AUC (0.025--0.110; $p=0.0024$). Tong9, Tong8, and Tong30 achieved AUCs of 0.767, 0.758, and 0.741, respectively. The 11-candidate literature COP union achieved AUC 0.828 (0.764--0.887), with AP-COP10 higher by 0.066 (0.022--0.110; $p=0.0024$). These were secondary sensitivity comparisons and were not part of the Holm-adjusted three-comparison family.

\section{Selection stability and feature-family competition}

Across 100 Full35 outer folds, the mean final model size was 5.92 descriptors and the median was 6. Pairwise Jaccard similarity had mean 0.620 and median 0.714. At least one stance-indexed AP-COP descriptor appeared in every outer-fold final model, with a mean of 3.73 AP-COP descriptors selected per fold. The descriptor-level retention pattern is summarized in Table~\ref{tab:apcop_retention}, with early-stance asymmetry and AP-COP position at 5--35\% among the most consistently selected candidates.

Within AP-COP10 evaluated alone, outer-fold retention was 100\% for position at 5--15\%, 100\% for position at 25--35\%, 29\% for position at 45--55\%, 96\% for position at 65--75\%, and 8\% for position at 85--95\%. Corresponding asymmetry retention was 100\%, 60\%, 9\%, 72\%, and 5\%. Position at 65--75\% had weak marginal separation ($g=-0.198$, 95\% CI $-0.490$ to 0.095; $p=0.192$), yet deleting it from AP-COP10 reduced AUC by 0.019 (95\% CI $-0.000$ to 0.041; $p=0.053$). These results illustrate that marginal separation and multivariable retention are not identical.

\begin{table}[ht]

\centering

\caption{Feature-family representation across Full35 final models.}

\scriptsize

\begin{tabularx}{\columnwidth}{Xcc}

\toprule

Family & Mean selected/fold & Folds represented\\

\midrule

Stance-indexed AP-COP & 3.73 & 100\%\\

Dynamic support & 0.93 & 93\%\\

Temporal/rhythm & 0.72 & 72\%\\

Broader COP rollover & 0.39 & 37\%\\

Vertical loading & 0.09 & 9\%\\

Other bilateral asymmetry & 0.06 & 6\%\\

\bottomrule

\end{tabularx}

\end{table}

\begin{table}[ht]

\centering

\caption{Outer-fold retention of the 10 stance-indexed AP-COP descriptors within Full35 final models.}\label{tab:apcop_retention}

\scriptsize

\setlength{\tabcolsep}{3pt}

\begin{tabularx}{\columnwidth}{Xcc}

\toprule

Feature & Family & Retention\\

\midrule

\texttt{cop\_landing\_asymmetry} & AP-COP asymmetry & 0.99\\

\texttt{cop\_pos\_25\_35} & AP-COP position & 0.99\\

\texttt{cop\_landing\_position} & AP-COP position & 0.90\\

\texttt{cop\_asym\_25\_35} & AP-COP asymmetry & 0.40\\

\texttt{cop\_pos\_65\_75} & AP-COP position & 0.23\\

\texttt{cop\_midstance\_position} & AP-COP position & 0.15\\

\texttt{cop\_asym\_65\_75} & AP-COP asymmetry & 0.05\\

\texttt{cop\_terminal\_position} & AP-COP position & 0.01\\

\texttt{cop\_asym\_45\_55} & AP-COP asymmetry & 0.01\\

\texttt{cop\_asym\_85\_95} & AP-COP asymmetry & 0.00\\

\bottomrule

\end{tabularx}

\end{table}

\section{Complementary Non-AP-COP25 representation}

\label{sec:nonapcop25}

Non-AP-COP25 was defined exactly as Full35 minus the 10 stance-indexed AP-COP candidates. It therefore contains the six temporal/rhythm, four dynamic-support, four non-AP-COP bilateral-asymmetry, six broader COP-rollover, and five vertical-loading descriptors listed in Table~\ref{tab:full35_dictionary}. This complementary representation used the same 20 repetitions $\times$ 5 outer folds, five-fold inner tuning, 100 stability-selection subsamples, 0.60 stability threshold, $|\rho|\geq0.85$ redundancy pruning, six-feature maximum, final class-balanced L2 logistic regression, and participant-level probability averaging used for the primary engineered representations.

\begin{table}[ht]

\centering

\caption{Performance of the complementary Non-AP-COP25 representation under the common nested evaluation.}

\label{tab:nonapcop25_performance}

\scriptsize

\setlength{\tabcolsep}{2.4pt}

\begin{tabularx}{\columnwidth}{Xccccc}

\toprule

Representation & Candidates & AUC (95\% CI) & BAcc & Sens. & Spec.\\

\midrule

Full35 & 35 & 0.908 (0.862--0.948) & 0.817 & 0.828 & 0.806\\

AP-COP10 & 10 & 0.894 (0.843--0.937) & 0.828 & 0.796 & 0.861\\

Non-AP-COP25 & 25 & 0.856 (0.796--0.910) & 0.751 & 0.753 & 0.750\\

\bottomrule

\end{tabularx}

\end{table}

Matched participant-level bootstrap comparisons gave Full35 minus Non-AP-COP25 $\Delta$AUC $=0.051$ (95\% CI 0.015--0.089; $p=0.0028$), AP-COP10 minus Non-AP-COP25 $\Delta$AUC $=0.038$ ($-0.011$ to 0.090; $p=0.134$), and Full35 minus AP-COP10 $\Delta$AUC $=0.013$ ($-0.015$ to 0.043; $p=0.361$). Thus, removing AP-COP10 from Full35 produced a statistically supported loss, whereas adding the complementary descriptors to AP-COP10 produced only a small, unsupported increase. These contrasts are secondary nested representation analyses and were not included in the Holm-adjusted literature-comparison family.

Non-AP-COP25 final models selected a mean of 5.91 descriptors per outer fold (median 6; range 3--6), and 94 of 100 outer folds reached the six-feature cap. The most frequently retained candidates are shown in Table~\ref{tab:nonapcop25_retention}. The similar final-model size indicates that the lower stand-alone AUC was not attributable simply to systematically smaller selected models.

\begin{table}[ht]

\centering

\caption{Most frequently retained descriptors within the stand-alone Non-AP-COP25 representation.}

\label{tab:nonapcop25_retention}

\scriptsize

\setlength{\tabcolsep}{3pt}

\begin{tabularx}{\columnwidth}{Xc}

\toprule

Feature & Outer-fold retention\\

\midrule

Step-interval MAD ratio & 0.98\\

Double-support proportion & 0.96\\

COP AP range & 0.94\\

COP early slope & 0.81\\

Peak-force asymmetry & 0.72\\

Normalized loading rate & 0.66\\

Stance-ratio asymmetry & 0.24\\

Median COP speed & 0.22\\

\bottomrule

\end{tabularx}

\end{table}

\section{Secondary robustness analyses}

\subsection{Feature-budget and final-feature-cap sensitivity}

\begin{table}[ht]

\centering

\caption{Full35 compactness sensitivity.}

\scriptsize

\setlength{\tabcolsep}{3pt}

\begin{tabularx}{\columnwidth}{Xccc}

\toprule

Analysis & Setting & AUC & 95\% CI\\

\midrule

Budget & 1 & 0.802 & 0.734--0.866\\

Budget & 2 & 0.865 & 0.810--0.914\\

Budget & 3 & 0.885 & 0.835--0.931\\

Budget & 4 & 0.901 & 0.855--0.941\\

Budget & 5 & 0.906 & 0.861--0.947\\

Budget & 6 & 0.908 & 0.862--0.947\\

Cap & 4 & 0.901 & 0.855--0.943\\

Cap & 6 & 0.908 & 0.862--0.948\\

Cap & 8 & 0.908 & 0.862--0.947\\

Cap & 10 & 0.905 & 0.859--0.944\\

Cap & 12 & 0.903 & 0.854--0.943\\

\bottomrule

\end{tabularx}

\end{table}

The six-descriptor budget exceeded the three-descriptor budget by $\Delta$AUC 0.022 (95\% CI 0.002--0.045; $p=0.033$), while gains over four and five descriptors were not supported. The larger final caps did not improve on six.

\subsection{Classifier-family sensitivity}

\begin{table}[ht]

\centering

\caption{Classifier-family sensitivity using the same fold-specific Full35 feature sets.}

\scriptsize

\setlength{\tabcolsep}{2.5pt}

\begin{tabularx}{\columnwidth}{Xcccc}

\toprule

Classifier & AUC & BAcc & Sens. & Spec.\\

\midrule

Histogram gradient boosting & 0.913 & 0.851 & 0.882 & 0.819\\

Logistic regression & 0.908 & 0.817 & 0.828 & 0.806\\

RBF-SVM & 0.899 & 0.803 & 0.828 & 0.778\\

Random forest & 0.898 & 0.817 & 0.871 & 0.764\\

\bottomrule

\end{tabularx}

\end{table}

Histogram gradient boosting produced the highest descriptive AUC (0.913), followed by logistic regression (0.908), RBF-SVM (0.899), and random forest (0.898). Calibration slopes were 1.01 for logistic regression and 0.73 for histogram gradient boosting. No paired inferential claim is made for the classifier-family AUC differences in this supplement.

\subsection{Sensor perturbation and descriptor stability}

\begin{table}[ht]

\centering

\caption{Full35 performance under simulated sensor perturbations.}

\scriptsize

\setlength{\tabcolsep}{2.5pt}

\begin{tabularx}{\columnwidth}{Xccc}

\toprule

Condition & AUC & BAcc & AUC loss\\

\midrule

Clean & 0.908 & 0.817 & ref.\\

5\% gain SD & 0.907 & 0.812 & 0.001\\

10\% gain SD & 0.902 & 0.814 & 0.006\\

20\% gain SD & 0.881 & 0.789 & 0.027\\

One-sensor dropout & 0.858 & 0.787 & 0.050\\

Two-sensor dropout & 0.802 & 0.738 & 0.106\\

\bottomrule

\end{tabularx}

\end{table}

Under 10\% gain variation, the five AP-COP position descriptors retained median Spearman agreement ranging from 0.985 to 0.996 with their clean values, whereas asymmetry descriptors ranged from 0.865 to 0.921. Under one-sensor dropout, position descriptors ranged from 0.831 to 0.966 and asymmetry descriptors from 0.544 to 0.618. The lowest asymmetry stability occurred at 65--75\% stance ($\rho=0.544$), while early-stance asymmetry had $\rho=0.618$; the terminal-stance position descriptor was the least stable position feature ($\rho=0.831$).

\begin{figure}[ht]

\centering

\includegraphics[width=0.94\columnwidth]{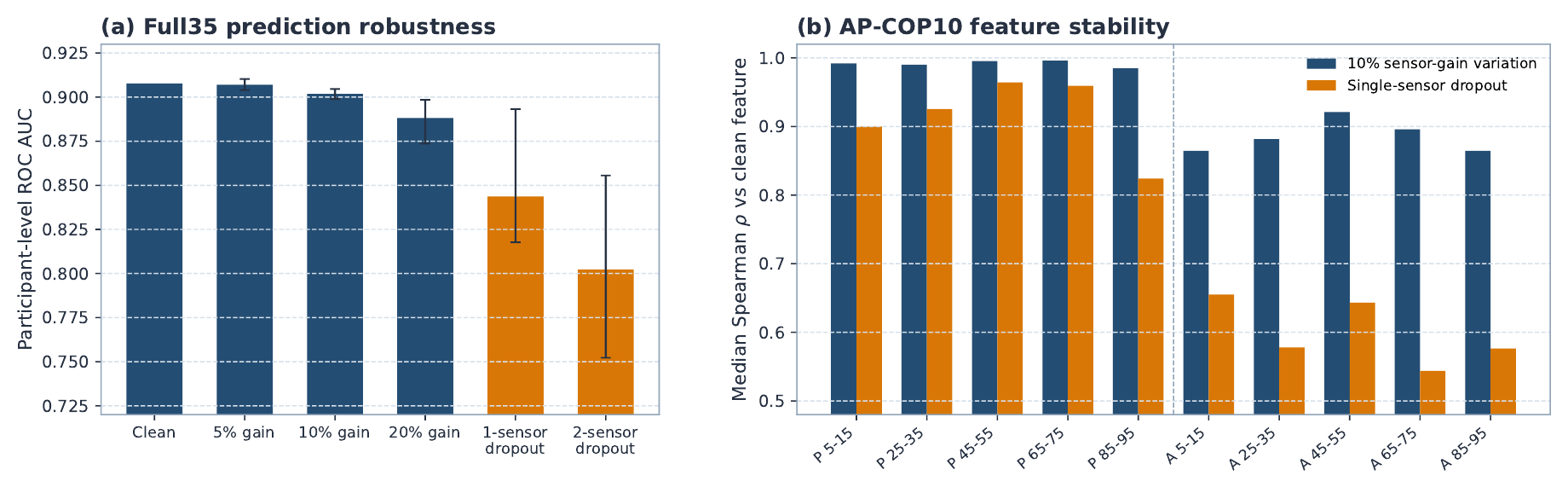}

\caption{Simulated sensor robustness and AP-COP descriptor stability. Left: Full35 performance under sensor perturbations. Right: AP-COP descriptor stability under sensor perturbations.}

\label{fig:sensor_robustness}

\end{figure}

\subsection{Asymmetry and covariate sensitivity}

\begin{table*}[t]

\centering

\caption{Sensitivity and attribution analyses for AP-COP10. AUC differences are reported for the listed representation minus the stated comparator.}

\label{tab:sensitivity_attribution}

\scriptsize

\setlength{\tabcolsep}{4pt}

\renewcommand{\arraystretch}{1.06}

\begin{tabularx}{0.97\textwidth}{>{\raggedright\arraybackslash}p{3.0cm} >{\raggedright\arraybackslash}p{5.0cm} c X}

\toprule

Analysis & Representation & AUC & Relevant paired comparison\\

\midrule

\multicolumn{4}{l}{\textit{Asymmetry-definition sensitivity}}\\

& Primary relative asymmetry (AP-COP10) & 0.894 & Reference.\\

& Absolute bilateral AP-COP difference & 0.896 & $\Delta$AUC $=+0.002$ vs primary (95\% CI $-0.005$ to 0.008; $p=0.539$).\\

& AP-range-normalized bilateral difference & 0.894 & $\Delta$AUC $=-0.000$ vs primary (95\% CI $-0.009$ to 0.008; $p=0.988$).\\

\addlinespace[2pt]

\multicolumn{4}{l}{\textit{Representation attribution}}\\

& Five AP-COP positions only & 0.857 & Position-only reference.\\

& Five positions + 5--15\% asymmetry & 0.900 & $\Delta$AUC $=+0.043$ vs positions only (0.009--0.079; $p=0.013$); $+0.006$ vs AP-COP10 ($-0.005$ to 0.017; $p=0.313$).\\

& Five asymmetry descriptors only & 0.746 & Descriptive secondary result.\\

\addlinespace[2pt]

\multicolumn{4}{l}{\textit{Walking-speed predictive sensitivity}}\\

& Walking speed only & 0.772 & Speed reference.\\

& Age + walking speed & 0.767 & Age-and-speed reference.\\

& AP-COP10 + speed & 0.919 & $\Delta$AUC $=+0.147$ vs speed only (0.085--0.215; $p<0.001$); $+0.025$ vs AP-COP10 ($-0.001$ to 0.053; $p=0.060$).\\

& AP-COP10 + age + speed & 0.922 & $\Delta$AUC $=+0.155$ vs age + speed (0.091--0.221; $p<0.001$).\\

\bottomrule

\end{tabularx}

\end{table*}

The asymmetry-definition sensitivities address coordinate-origin dependence of the primary relative index. The attribution and covariate analyses are secondary analyses and did not redefine AP-COP10. The speed analyses are predictive sensitivities rather than causal covariate adjustment.

\subsection{Source-study holdout and study-wise calibration}

Leave-one-source-study-out AUCs were 0.837, 0.937, and 0.946 for AP-COP10 and 0.807, 0.932, and 0.867 for Full35 when Ga, Ju, and Si were held out, respectively. The pooled values were 0.907 for AP-COP10 and 0.872 for Full35. This analysis remains internal to GaitPDB and is not independent external validation.

\begin{table}[ht]

\centering

\caption{Leave-one-source-study-out AUC with 95\% bootstrap confidence intervals.}

\scriptsize

\setlength{\tabcolsep}{2.4pt}

\begin{tabularx}{\columnwidth}{>{\raggedright\arraybackslash}p{1.25cm} *{4}{>{\centering\arraybackslash}X}}

\toprule

Rep. & Pooled & Ga & Ju & Si\\

\midrule

Full35 & 0.872\newline(0.813--0.921) & 0.807\newline(0.670--0.920) & 0.932\newline(0.859--0.983) & 0.867\newline(0.767--0.948)\\

AP-COP10 & 0.907\newline(0.860--0.947) & 0.837\newline(0.713--0.939) & 0.937\newline(0.865--0.988) & 0.946\newline(0.890--0.986)\\

\bottomrule

\end{tabularx}

\end{table}

Study-wise repeated-CV Full35 AUCs were 0.918 for Ga, 0.934 for Ju, and 0.888 for Si. The overall Full35 calibration slope was 1.01; study-wise slopes were 1.25, 1.15, and 0.87.

\section{Reproducibility configuration}

\begin{table}[ht]

\centering

\caption{Fixed configuration values used in the verified analyses.}

\scriptsize

\setlength{\tabcolsep}{3pt}

\begin{tabularx}{\columnwidth}{>{\raggedright\arraybackslash}p{2.7cm}X}

\toprule

Setting & Value\\

\midrule

Participants & 165 (93 PD, 72 controls)\\

Outer evaluation & 20 repetitions $\times$ 5 folds\\

Outer stratification & Study $\times$ Diagnosis\\

Stability subsamples & 100 per outer fold\\

Stability threshold & 60\% nonzero-selection frequency\\

Redundancy threshold & $|\rho|\geq0.85$ (Spearman)\\

Maximum final features & At most 6 for all representations (candidate count permitting)\\

Final classifier & Class-balanced L2 logistic regression\\

Classifier parameter & $C=1$, \texttt{liblinear}\\

Inner elastic-net grid & $C$: $10^{-2}$--$10^1$ (9 values); l1 ratio: 0.25, 0.50, 0.75, 1.00\\

Complementary representation & Non-AP-COP25 = Full35 minus AP-COP10\\

Bootstrap resamples & 5,000\\

Sensor replicates & 5 per perturbation condition\\

Software & Python 3.10.19; NumPy 1.26.4; pandas 2.3.3; Linux\\

\bottomrule

\end{tabularx}

\end{table}

The exact feature identifiers in Tables~\ref{tab:apcop10} and \ref{tab:full35_dictionary} are the column names used in the final feature matrices and analysis notebooks. Participant-level predictions, paired contrasts, fold-level selections, selection-stability summaries, and supporting numerical outputs are retained by the authors and can be made available from the authors upon reasonable request. The numerical results required to interpret the manuscript are reported in the main text and this supplement.


\begin{thebibliography}{99}

\bibitem{mirelman2019}
A. Mirelman, P. Bonato, R. Camicioli, T.D. Ellis, N. Giladi, J.L. Hamilton, C.J. Hass, J.M. Hausdorff, E. Pelosin, Q.J. Almeida, Gait impairments in Parkinson's disease, \textit{Lancet Neurology} 18 (7) (2019) 697--708. doi:10.1016/S1474-4422(19)30044-4.

\bibitem{bouca2020}
R. Bou\c{c}a-Machado, C. Jalles, D. Guerreiro, F. Pona-Ferreira, D. Branco, T. Guerreiro, R. Matias, J.J. Ferreira, Gait kinematic parameters in Parkinson's disease: a systematic review, \textit{Journal of Parkinson's Disease} 10 (3) (2020) 843--853. doi:10.3233/JPD-201969.

\bibitem{bloem2021}
B.R. Bloem, M.S. Okun, C. Klein, Parkinson's disease, \textit{The Lancet} 397 (10291) (2021) 2284--2303. doi:10.1016/S0140-6736(21)00218-X.

\bibitem{russo2025}
M. Russo, M. Amboni, N. Pisani, A. Volzone, D. Calderone, P. Barone, F. Amato, C. Ricciardi, M. Romano, Biomechanics parameters of gait analysis to characterize Parkinson's disease: a scoping review, \textit{Sensors} 25 (2) (2025) 338. doi:10.3390/s25020338.

\bibitem{mancini2025}
M. Mancini, M. Afshari, Q. Almeida, et al., Digital gait biomarkers in Parkinson's disease: susceptibility/risk, progression, response to exercise, and prognosis, \textit{npj Parkinson's Disease} 11 (1) (2025) 51. doi:10.1038/s41531-025-00897-1.

\bibitem{zhang2023insole}
Z. Zhang, Y. Dai, Z. Xu, N. Grimaldi, J. Wang, M. Zhao, R. Pang, Y. Sun, S. Gao, B. Hu, Insole systems for disease diagnosis and rehabilitation: a review, \textit{Biosensors} 13 (8) (2023) 833. doi:10.3390/bios13080833.

\bibitem{wang2026survey}
X. Wang, Z. Zhao, L. Lin, F. Li, F. Qi, X. Wang, L. Han, A comprehensive survey on diagnosis and assessment of Parkinson's disease via plantar pressure analysis, \textit{npj Parkinson's Disease} (2026), online ahead of print. doi:10.1038/s41531-026-01416-6.

\bibitem{hausdorff2008}
J.M. Hausdorff, Gait in Parkinson's Disease, PhysioNet, version 1.0.0 (2008). doi:10.13026/C24H3N.

\bibitem{han1999}
T.R. Han, N.J. Paik, M.S. Im, Quantification of the path of center of pressure (COP) using an F-scan in-shoe transducer, \textit{Gait \& Posture} 10 (3) (1999) 248--254. doi:10.1016/S0966-6362(99)00040-5.

\bibitem{lugade2014}
V. Lugade, K. Kaufman, Center of pressure trajectory during gait: a comparison of four foot positions, \textit{Gait \& Posture} 40 (4) (2014) 719--722. doi:10.1016/j.gaitpost.2014.07.001.

\bibitem{shin2020}
C. Shin, T.-B. Ahn, Asymmetric dynamic center-of-pressure in Parkinson's disease, \textit{Journal of the Neurological Sciences} 408 (2020) 116559. doi:10.1016/j.jns.2019.116559.

\bibitem{zhang2024}
X. Zhang, Y. Li, P. Wang, Q. Zhao, Temporal and spatial characteristics of plantar pressure center trajectory for identifying early Parkinson's disease gait, \textit{Parkinsonism \& Related Disorders} 124 (2024) 106998. doi:10.1016/j.parkreldis.2024.106998.

\bibitem{pataky2014}
T.C. Pataky, M.A. Robinson, J. Vanrenterghem, R. Savage, K.T. Bates, R.H. Crompton, Vector field statistics for objective center-of-pressure trajectory analysis during gait, with evidence of scalar sensitivity to small coordinate system rotations, \textit{Gait \& Posture} 40 (1) (2014) 255--258. doi:10.1016/j.gaitpost.2014.01.023.

\bibitem{chiu2013speed}
M.-C. Chiu, H.-C. Wu, L.-Y. Chang, Gait speed and gender effects on center of pressure progression during normal walking, \textit{Gait \& Posture} 37 (1) (2013) 43--48. doi:10.1016/j.gaitpost.2012.05.030.

\bibitem{buldt2018}
A.K. Buldt, S. Forghany, K.B. Landorf, G.S. Murley, P. Levinger, H.B. Menz, Centre of pressure characteristics in normal, planus and cavus feet, \textit{Journal of Foot and Ankle Research} 11 (2018) 3. doi:10.1186/s13047-018-0245-6.

\bibitem{nieuwboer1999}
A. Nieuwboer, W. De Weerdt, R. Dom, L. Peeraer, E. Lesaffre, F. Hilde, B. Baunach, Plantar force distribution in Parkinsonian gait: a comparison between patients and age-matched control subjects, \textit{Scandinavian Journal of Rehabilitation Medicine} 31 (3) (1999) 185--192. doi:10.1080/003655099444533.

\bibitem{hausdorff1998}
J.M. Hausdorff, M.E. Cudkowicz, R. Firtion, J.Y. Wei, A.L. Goldberger, Gait variability and basal ganglia disorders: stride-to-stride variations of gait cycle timing in Parkinson's disease and Huntington's disease, \textit{Movement Disorders} 13 (3) (1998) 428--437. doi:10.1002/mds.870130310.

\bibitem{chiu2013elderly}
M.-C. Chiu, H.-C. Wu, L.-Y. Chang, M.-H. Wu, Center of pressure progression characteristics under the plantar region for elderly adults, \textit{Gait \& Posture} 37 (3) (2013) 408--412. doi:10.1016/j.gaitpost.2012.08.010.

\bibitem{fadil2021}
R. Fadil, A. Huether, R. Brunnemer, A.P. Blaber, J.-S. Lou, K. Tavakolian, Early detection of Parkinson's disease using center of pressure data and machine learning, in: \textit{2021 43rd Annual International Conference of the IEEE Engineering in Medicine \& Biology Society (EMBC)}, 2021, pp. 2433--2436. doi:10.1109/EMBC46164.2021.9630451.

\bibitem{sun2023}
Y. Sun, Y. Cheng, Y. You, Y. Wang, Z. Zhu, Y. Yu, J. Han, J. Wu, N. Yu, A novel plantar pressure analysis method to signify gait dynamics in Parkinson's disease, \textit{Mathematical Biosciences and Engineering} 20 (8) (2023) 13474--13490. doi:10.3934/mbe.2023601.

\bibitem{nanayakkara2025}
T. Nanayakkara, H.M.K.K.M.B. Herath, H.S. Malekroodi, N. Madusanka, M. Yi, B.-I. Lee, Multi-domain CoP feature analysis of functional mobility for Parkinson's disease detection using wearable pressure insoles, \textit{Sensors} 25 (18) (2025) 5859. doi:10.3390/s25185859.

\bibitem{alam2017}
M.N. Alam, A. Garg, T.T.K. Munia, R. Fazel-Rezai, K. Tavakolian, Vertical ground reaction force marker for Parkinson's disease, \textit{PLOS ONE} 12 (5) (2017) e0175951. doi:10.1371/journal.pone.0175951.

\bibitem{jin2026}
L. Jin, Identifying Parkinson's Disease from Gait Biomechanics Using a Participant-Level Machine Learning Analysis Pipeline, \textit{Applied Sciences} 16 (13) (2026) 6296. doi:10.3390/app16136296.

\bibitem{tong2021}
J. Tong, J. Zhang, E. Dong, S. Du, Severity Classification of Parkinson's Disease Based on Permutation-Variable Importance and Persistent Entropy, \textit{Applied Sciences} 11 (4) (2021) 1834. doi:10.3390/app11041834.

\bibitem{zeng2016}
W. Zeng, F. Liu, Q. Wang, Y. Wang, L. Ma, Y. Zhang, Parkinson's disease classification using gait analysis via deterministic learning, \textit{Neuroscience Letters} 633 (2016) 268--278. doi:10.1016/j.neulet.2016.09.043.

\bibitem{khoury2019}
N. Khoury, F. Attal, Y. Amirat, L. Oukhellou, S. Mohammed, Data-driven based approach to aid Parkinson's disease diagnosis, \textit{Sensors} 19 (2) (2019) 242. doi:10.3390/s19020242.

\bibitem{farashi2020}
S. Farashi, Distinguishing between Parkinson's disease patients and healthy individuals using a comprehensive set of time, frequency and time-frequency features extracted from vertical ground reaction force data, \textit{Biomedical Signal Processing and Control} 62 (2020) 102132. doi:10.1016/j.bspc.2020.102132.

\bibitem{balaji2020}
E. Balaji, D. Brindha, R. Balakrishnan, Supervised machine learning based gait classification system for early detection and stage classification of Parkinson's disease, \textit{Applied Soft Computing} 94 (2020) 106494. doi:10.1016/j.asoc.2020.106494.

\bibitem{farashi2021}
S. Farashi, Analysis of the stance phase of the gait cycle in Parkinson's disease and its potency for Parkinson's disease discrimination, \textit{Journal of Biomechanics} 129 (2021) 110818. doi:10.1016/j.jbiomech.2021.110818.

\bibitem{elmaachi2020}
I. El Maachi, G.-A. Bilodeau, W. Bouachir, Deep 1D-Convnet for accurate Parkinson disease detection and severity prediction from gait, \textit{Expert Systems with Applications} 143 (2020) 113075. doi:10.1016/j.eswa.2019.113075.

\bibitem{choi2026}
M. Choi, J. Jo, J. Jeong, Digital gait biomarkers for Parkinson's disease: subject-wise validated explainable AI framework using vertical ground reaction force signals, \textit{Bioengineering} 13 (3) (2026) 360. doi:10.3390/bioengineering13030360.

\bibitem{wang2026graph}
X. Wang, X. Xu, Z. Zhao, F. Li, F. Qi, S. Liang, VGRF signal-based gait analysis for Parkinson's disease detection: a multi-scale directed graph neural network approach, \textit{IEEE Journal of Biomedical and Health Informatics} 30 (1) (2026) 100--112. doi:10.1109/JBHI.2025.3589772.

\bibitem{savitzky1964}
A. Savitzky, M.J.E. Golay, Smoothing and differentiation of data by simplified least squares procedures, \textit{Analytical Chemistry} 36 (8) (1964) 1627--1639. doi:10.1021/ac60214a047.

\bibitem{robinson1987}
R.O. Robinson, W. Herzog, B.M. Nigg, Use of force platform variables to quantify the effects of chiropractic manipulation on gait symmetry, \textit{Journal of Manipulative and Physiological Therapeutics} 10 (4) (1987) 172--176. PMID:2958572.

\bibitem{viteckova2018}
S. Viteckova, P. Kutilek, Z. Svoboda, R. Krupicka, J. Kauler, Z. Szabo, Gait symmetry measures: A review of current and prospective methods, \textit{Biomedical Signal Processing and Control} 42 (2018) 89--100. doi:10.1016/j.bspc.2018.01.013.

\bibitem{varma2006}
S. Varma, R. Simon, Bias in error estimation when using cross-validation for model selection, \textit{BMC Bioinformatics} 7 (2006) 91. doi:10.1186/1471-2105-7-91.

\bibitem{cawley2010}
G.C. Cawley, N.L.C. Talbot, On over-fitting in model selection and subsequent selection bias in performance evaluation, \textit{Journal of Machine Learning Research} 11 (2010) 2079--2107.

\bibitem{zou2005}
H. Zou, T. Hastie, Regularization and variable selection via the elastic net, \textit{Journal of the Royal Statistical Society: Series B} 67 (2) (2005) 301--320. doi:10.1111/j.1467-9868.2005.00503.x.

\bibitem{meinshausen2010}
N. Meinshausen, P. B\"uhlmann, Stability selection, \textit{Journal of the Royal Statistical Society: Series B} 72 (4) (2010) 417--473. doi:10.1111/j.1467-9868.2010.00740.x.

\bibitem{holm1979}
S. Holm, A simple sequentially rejective multiple test procedure, \textit{Scandinavian Journal of Statistics} 6 (2) (1979) 65--70.

\bibitem{geurts2011}
A.C.H. Geurts, T.A. Boonstra, N.C. Voermans, M.G. Diender, V. Weerdesteyn, B.R. Bloem, Assessment of postural asymmetry in mild to moderate Parkinson's disease, \textit{Gait \& Posture} 33 (1) (2011) 143--145. doi:10.1016/j.gaitpost.2010.09.018.

\bibitem{beretta2015}
V.S. Beretta, L.T.B. Gobbi, E. Lirani-Silva, L. Simieli, D. Orcioli-Silva, F.A. Barbieri, Challenging postural tasks increase asymmetry in patients with Parkinson's disease, \textit{PLOS ONE} 10 (9) (2015) e0137722. doi:10.1371/journal.pone.0137722.

\bibitem{frenkel2005}
S. Frenkel-Toledo, N. Giladi, C. Peretz, T. Herman, L. Gruendlinger, J.M. Hausdorff, Effect of gait speed on gait rhythmicity in Parkinson's disease: variability of stride time and swing time respond differently, \textit{Journal of NeuroEngineering and Rehabilitation} 2 (2005) 23. doi:10.1186/1743-0003-2-23.

\end{thebibliography}
\end{document}